%% file: main.tex
\documentclass[acmtog,nonacm]{acmart}

\input{preamble}  
\input{sections/commands}

\setcitestyle{numbers,comma}
\usepackage{numprint}
\usepackage[ruled]{algorithm2e} %
\usepackage{float}
\usepackage{caption}
\usepackage{subcaption}
\usepackage{placeins}
\usepackage{wrapfig}

\SetAlFnt{\small}
\SetAlCapFnt{\small}
\SetAlCapNameFnt{\small}
\SetAlCapHSkip{0pt}

\acmJournal{TOG}

\begin{document}

\author{Amirhossein Alimohammadi$^{*}$}
\affiliation{%
  \institution{Simon Fraser University}
  \country{Canada}
}
\author{Aryan Mikaeili$^{*}$}
\affiliation{%
  \institution{Simon Fraser University}
  \country{Canada}
}
\author{Sauradip Nag}
\affiliation{%
  \institution{Simon Fraser University}
  \country{Canada}
}
\author{Negar Hassanpour}
\affiliation{%
  \institution{Huawei}
  \country{Canada}
}
\author{Andrea Tagliasacchi}
\affiliation{%
  \institution{Simon Fraser University,}
  \institution{University of Toronto,}
  \institution{Google Deepmind}
  \country{Canada}
}
\author{Ali Mahdavi-Amiri}
\affiliation{%
  \institution{Simon Fraser University}
  \country{Canada}
}

\title{\textcolor{title_purple}{Cora}: \textcolor{title_purple}{Cor}respondence-\textcolor{title_purple}{a}ware image editing using few step diffusion} 



\begin{abstract}
  \input{sections/0_abstract}
  \footnote{$^*$ Authors have contributed equally}
\end{abstract}

%
%


%
%


\begin{teaserfigure}
  \includegraphics[width=\textwidth]{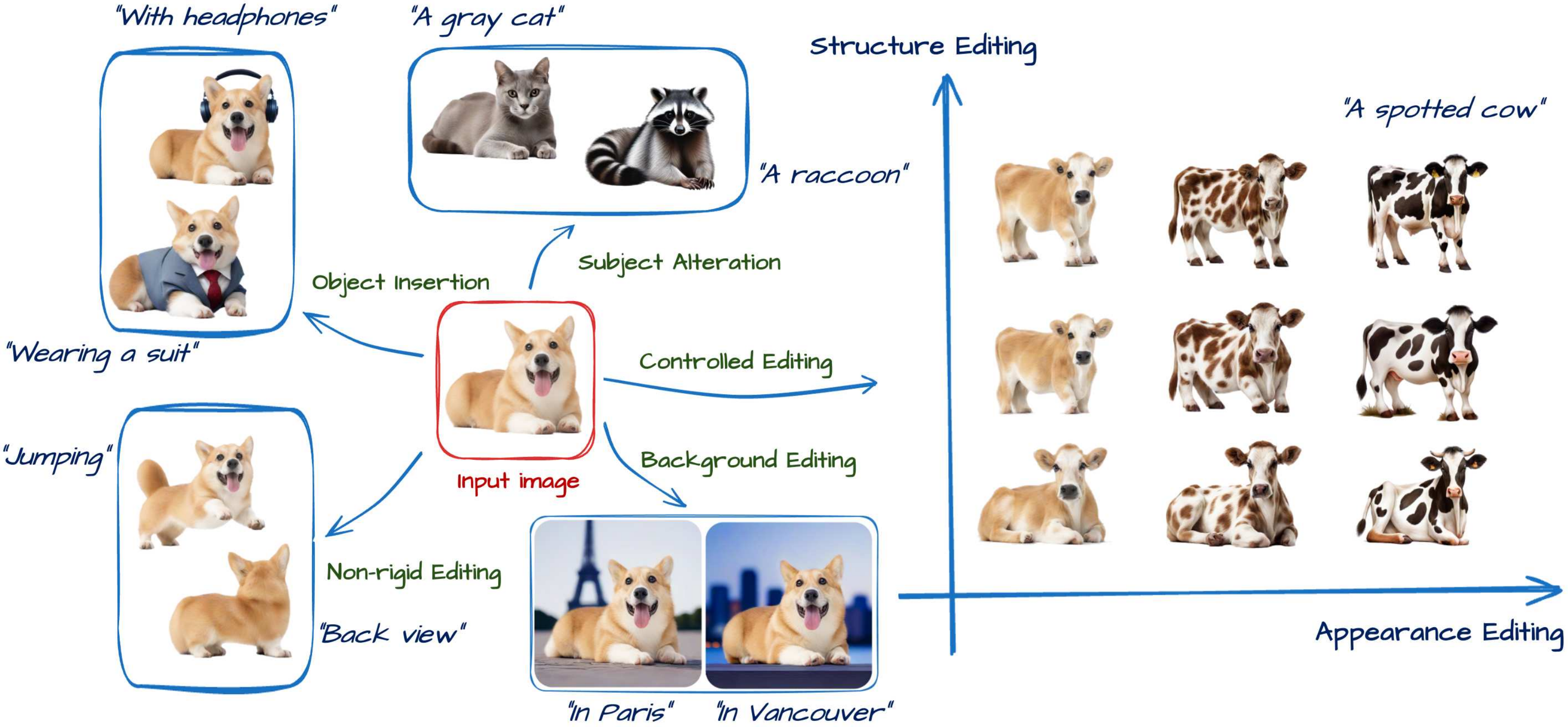}
  \caption{\papernospace{} supports diverse edits, including object insertion, subject and background changes, and non-rigid deformations (e.g., jumping). Our novel correspondence-aware method provides strong control and flexibility for both appearance and structure editing.}
  \Description{}
  \label{fig:teaser}
\end{teaserfigure}

\maketitle

\input{sections/1_introduction}

\input{sections/2_related_works}


\input{sections/3_method}

\input{sections/4_experiments}

\input{sections/5_conclusion}

\bibliographystyle{ACM-Reference-Format}
\bibliography{main}
\newpage

\begin{figure*}
  \includegraphics[width=\textwidth]{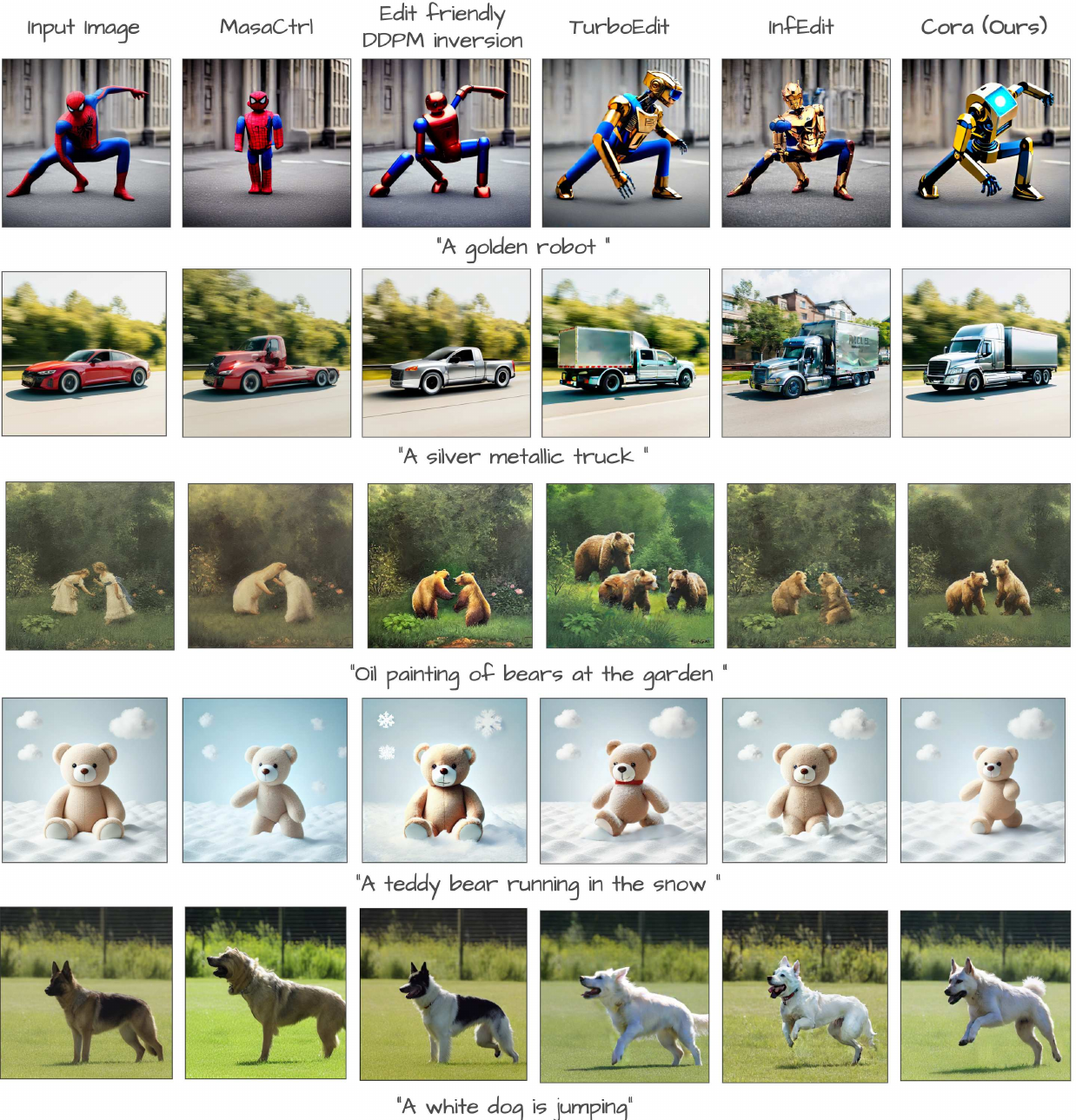}
  \caption{\textbf{Qualitative comparison}. We compare the visual quality of our edits with several few-step and multi-step frameworks. our method strikes the perfect balance between respecting the editing text-prompt and preserving the content of the input image.}
  \label{fig:comparison}
\end{figure*}

\begin{figure*}
  \includegraphics[width=0.9\textwidth]{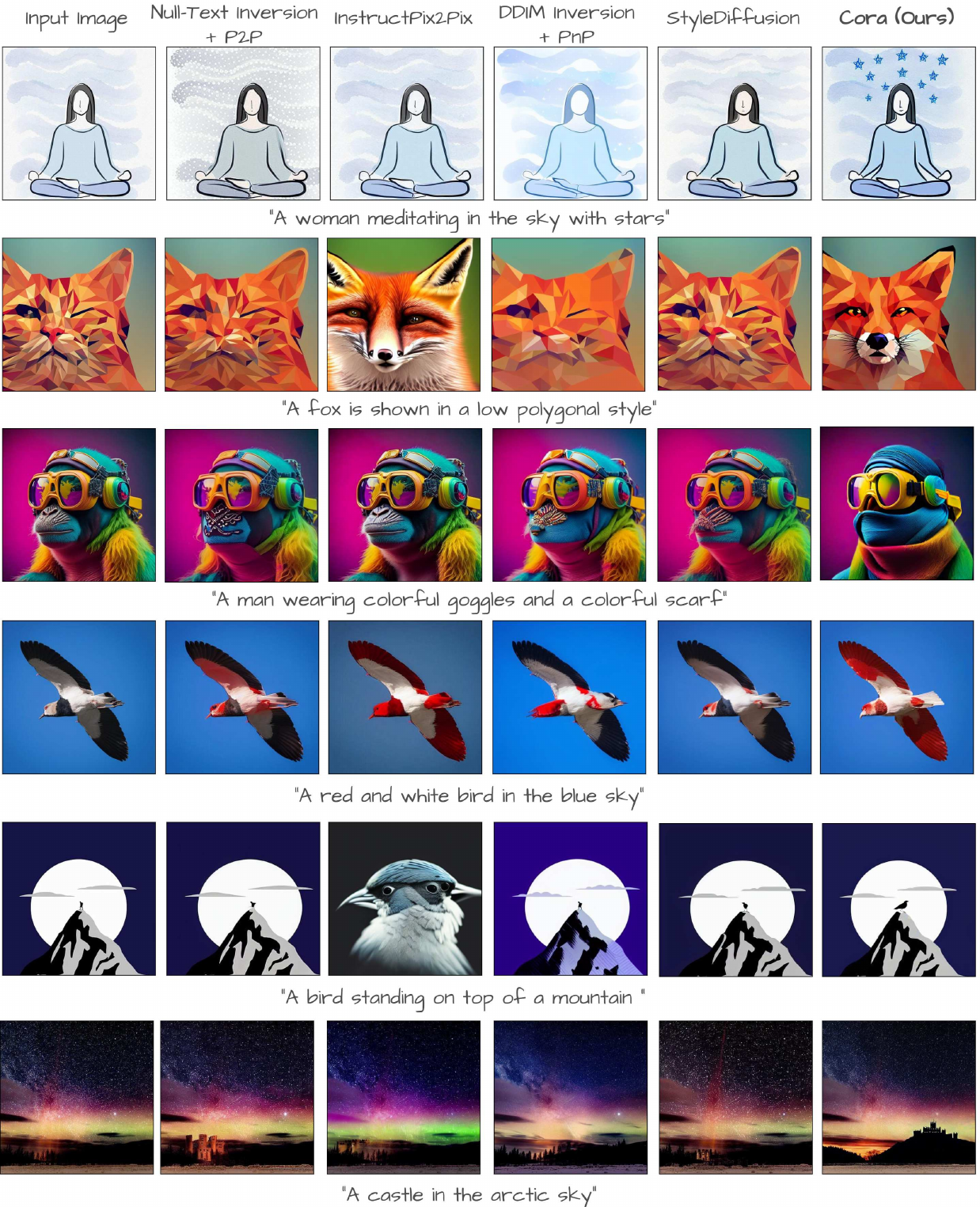}
  \caption{\textbf{Additional Qualitative Comparison.} We compare our method against several editing frameworks, including Null-Text Inversion~\cite{mokady2022null}, InstructPix2Pix~\cite{brooks2022instructpix2pix}, Plug-and-Play~\cite{tumanyan2022plugandplay}, and StyleDiffusion~\cite{li2023stylediffusion}.}
  \label{fig:comparison_extra}
\end{figure*}

\clearpage
\input{sections/supplementary}

\end{document}

%% file: preamble.tex
\usepackage{color}
\usepackage{graphicx}
\usepackage{caption}
\usepackage{wrapfig}
\usepackage{caption}
\usepackage{wrapfig}
\usepackage{wrapfig}

\definecolor{title_purple}{rgb}{0.65,0.1,0.65}

\definecolor{green}{rgb}{0, 0.5, 0}
\definecolor{orange}{rgb}{0.8, 0.6, 0.2}
\definecolor{red}{rgb}{1.0, 0.0, 0.0}
\definecolor{teal}{rgb}{0.0, 0.4, 0.4}
\definecolor{purple}{rgb}{0.65,0.0,0.65}
\definecolor{saffron}{rgb}{0.95,0.75,0.2}
\definecolor{turquoise}{rgb}{0.0,0.5,0.5}
\definecolor{black}{rgb}{0.0, 0.0, 0.0}
\definecolor{gray}{rgb}{0.5, 0.5, 0.5}

\newcommand{\papernospace}{Cora}
\newcommand{\paper}{Cora }

\definecolor{darkpink}{rgb}{0.561, 0.282, 0.427}
\definecolor{darkturquoise}{rgb}{0., 0.81, 0.822}

\makeatletter
\DeclareRobustCommand\onedot{\futurelet\@let@token\@onedot}

\makeatother

\makeatletter
\def\blfootnote{\xdef\@thefnmark{}\@footnotetext}
\makeatother

\DeclareMathOperator*{\argmin}{arg\,min}

%% file: sections/commands.tex
\usepackage{color}
\definecolor{green}{rgb}{0, 0.5, 0}
\definecolor{deepred}{rgb}{0.9, 0.15, 0.15}
\definecolor{orange}{rgb}{0.8, 0.6, 0.2}
\definecolor{red}{rgb}{1.0, 0.0, 0.0}
\definecolor{teal}{rgb}{0.0, 0.4, 0.4}
\definecolor{purple}{rgb}{0.65,0,0.65}
\definecolor{saffron}{rgb}{0.75,0.05,0.05}
\definecolor{turquoise}{rgb}{0.0,0.30,0.15}
\definecolor{black}{rgb}{0.0, 0.0, 0.0}
\definecolor{gray}{rgb}{0.5, 0.5, 0.5}


%% file: sections/0_abstract.tex
Image editing is an important task in computer graphics, vision, and VFX, with recent diffusion-based methods achieving fast and high-quality results. However, edits requiring significant structural changes, such as non-rigid deformations, object modifications, or content generation, remain challenging. Existing few step editing approaches produce artifacts such as irrelevant texture or struggle to preserve key attributes of the source image (e.g., pose). We introduce \paper, a novel editing framework that addresses these limitations by introducing correspondence-aware noise correction and interpolated attention maps. Our method aligns textures and structures between the source and target images through semantic correspondence, enabling accurate texture transfer while generating new content when necessary. \paper offers control over the balance between content generation and preservation. Extensive experiments demonstrate that, quantitatively and qualitatively, \paper excels in maintaining structure, textures, and identity across diverse edits, including pose changes, object addition, and texture refinements. User studies confirm that \paper delivers superior results, outperforming alternatives.

\noindent

\textbf{Project Page:} \href{https://cora-edit.github.io/}{\textcolor{title_purple}{cora-edit.github.io}}

%% file: sections/1_introduction.tex
\section{Introduction}
\label{sec:introduction}
Image editing is an important task in fields such as computer graphics, computer vision, and VFX. Recent diffusion-based, few-step image editing techniques have significantly improved this process, enabling fast and effective edits with impressive results across diverse scenarios~\cite{deutch2024turboedittextbasedimageediting,xu2023infedit, wu2024turboedit}.

\begin{figure}[ht]
    \centering
    \includegraphics[width=0.9\linewidth]{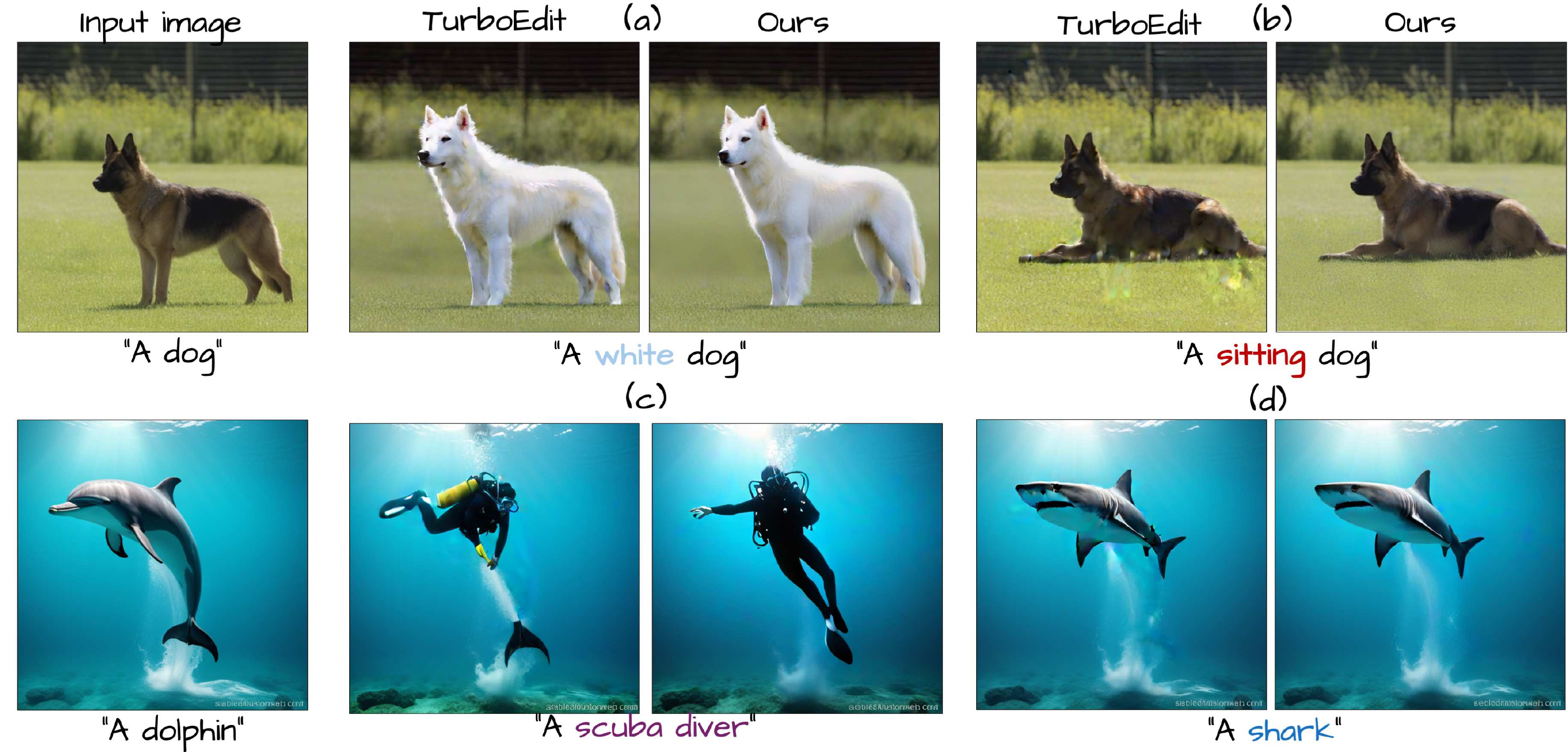}
    \caption{Comparison between TurboEdit~\cite{deutch2024turboedittextbasedimageediting} and our correspondence-aware editing approach. Due to misalignment between the source and target images, artifacts are visible in TurboEdit results, such as texture inconsistencies in \textbf{(a)}, silhouette artifacts in the legs and fins in \textbf{(b, d)}, and undesired elements in \textbf{(c)}. Please zoom in for a clearer view of these artifacts.
    }\label{fig:dog1}
\end{figure}

Despite these impressive advancements, edits that require structural changes that go beyond pixel color alteration (e.g., non-rigid edits, object change) remain a challenging task for diffusion models.
TurboEdit~\cite{deutch2024turboedittextbasedimageediting} that relies on noise \textit{correction} to perform edits, often produces artifacts and cannot necessarily preserve identity or important properties of the source image (e.g., pose); see~Fig.~\ref{fig:dog1}.
This is because these corrections do not account for the fact that the generated and source image may not any longer be pixel-aligned after the edit.
We resolve this shortcoming by introducing ``correspondence-aware'' noise corrections that connect source to target pixels by matching their diffusion features.

Edits involving significant deformation of the subject in the image~(Fig.~\ref{fig:dog1}), often require the generation of new parts, or the exposure of regions not present in the source image.
Some approaches that aim to respect the source for such edits primarily rely on the source image for texture information to maintain the subject's identity~\cite{Cao_2023_ICCV}.
While this strategy is somewhat effective, since they inject the intermediate features of the diffusion model from the source image into the self-attention modules~\cite{Cao_2023_ICCV}, their edits copy undesired texture from the source into regions of the target image with no clear correspondence~(Fig~\ref{fig:att_strategy}:b). 

One of our technical contributions is to combine the keys and values that carry texture information from both source and target.
This enables the network to generate content when needed while accurately copying textures when relevant information is available in the source image. 
However, simple methods for combining the source and target, such as concatenation, fail to achieve the desired results (see Sec.\ref{sec:spherical}).
We show that interpolating attention maps enhances performance while offering flexibility and control in both generating new content and preserving existing content.

To achieve the right textures while respecting the structure of the source image, it is also necessary to align attentions by establishing a semantic correspondence.
Therefore, we incorporated a correspondence technique (DIFT) into our method whenever source information is available.
This technique aligns the attention maps~(i.e., keys and values) of the source and target, enabling a more accurate and effective transfer of relevant textures.
In the early stages of generation, the model's output is primarily noise, making correspondence infeasible.
Therefore, in a four-step diffusion process, we initiate the correspondence process at the last two steps, where the image structure is established, but textures are still being refined.
To align the structure of the source and target images, we apply a permutation on queries that are obtained using a matching algorithm.
This alignment is performed in the first step of generation, as the image structure takes shape during this phase.

By combining these strategies, we present \papernospace, a novel editing method built upon a few-step text-to-image model, SDXL-Turbo~\cite{sauer2024fasthighresolutionimagesynthesis}.
We demonstrate that \paper delivers improved visual results for various edits, thanks to our innovative attention mixing and correspondence-aware techniques.
\paper not only excels at preserving the structure of the image and maintaining textures but also supports a wide range of edits, including non-rigid edits (e.g., pose changes), object addition and removal, and texture modifications; see Figure~\ref{fig:teaser}.
Furthermore, quantitative and qualitative experiments and user studies demonstrate the effectiveness of each component of our method and confirm that \paper surpasses other alternatives.

%% file: sections/2_related_works.tex
\begin{figure*}[ht]
    \centering
    \includegraphics[width=\linewidth]{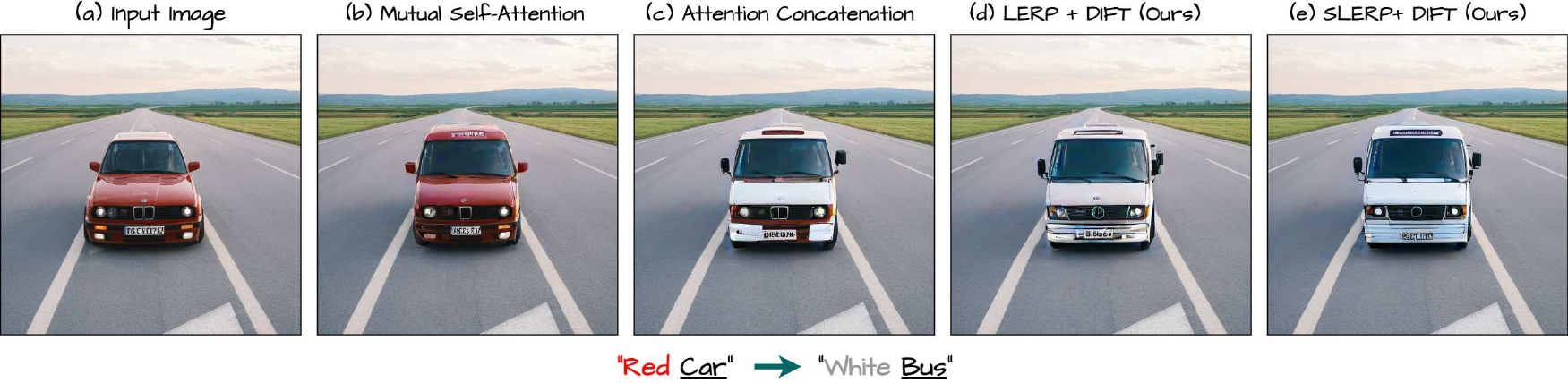}
    \caption{\textbf{Attention mixing strategies.} \textbf{(b):} Using only the source image's keys and values causes artifacts and misalignment with the edit text-prompt. \textbf{(c):} Concatenating keys and values leads to undesired appearance bleeding between source and target images. \textbf{(d, e):} Aligning and interpolating keys and values produces the best results, with \textit{slerp} providing more realistic and natural outcomes compared to \textit{lerp}.}
    \label{fig:att_strategy}
\end{figure*}

\section{Related Works}
\label{sec:related_works}


\subsubsection*{\textbf{Image editing with diffusion models}}
Diffusion-based text-to-image models are known for their ability to produce high-quality and diverse outputs~\cite{balaji2022eDiff-I, imagen2022saharia, rombach2021highresolution, ramesh2022dalle}.
These models start with Gaussian noise, and iteratively denoise it to generate the final image~\cite{sohl2015diffusion, ho2020ddpm}.
Beyond generation, they can also be leveraged as a powerful prior for visual content editing.
This is achieved by adding noise to an image and iteratively denoising it with a new text prompt~\cite{meng2022sdedit}, but this method often struggles to balance content preservation with prompt alignment.
To address these challenges, follow-up works tweaked the architecture, for example by including editing masks, to better preserve the content in the original image~\cite{avrahami2020blendeddiffusion, safaee2023clic, patashnik2023localizing, mirzaei2023watchyoursteps, hertz2023delta, Nam_2024_CVPR}.
Another approach involves training diffusion models on purpose-built datasets with images, instructional prompts, and ground truth edits~\cite{brooks2022instructpix2pix}, or fine-tuning models on the source image for improved content preservation~\cite{kawar2023imagic}.
However, since training these methods can be both time-consuming and resource-intensive, recent works have focused on leveraging the features of diffusion models for \textit{zero-shot} editing.

\subsubsection*{\textbf{Diffusion models' features}}
The features of diffusion models have been shown to contain rich spatial and semantic information.
These features can be used for tasks such as point correspondence~\cite{hedlin2023unsupervised, hedlin2024keypoints, tang2023emergent, luo2023dhf}, image and video segmentation~\cite{khani2023slime, alimohammadi2024smitesegmenttime}, and conditional image generation~\cite{luo2024readoutguidance, bhat2023loosecontrol}.
Some approaches leverage the cross-attention maps of diffusion models for localized text-based image editing~\cite{hertz2022prompt, mokady2022null}.
\citeN{tumanyan2023pnp} inject self-attention and intermediate features of the source image into the generation process of the target image for better content preservation. MasaCtrl~\cite{Cao_2023_ICCV} observes that using the keys and values of the self-attention modules from the source image allows copying its appearance into the target image while enabling non-rigid editing. However, as we will show, only using features from the source image is not sufficient for generating new content in the edited image.

Later works use this observation for style transfer~\cite{alaluf2023crossimage}, style-consistent generation~\cite{hertz2023StyleAligned}, image editing~\cite{koo2024flexiedit, lin2024ctrlx, patashnik2024qnerf} and video editing~\cite{tokenflow2023, qi2023fatezero}.
ConsiStory~\cite{tewel2024consistory} employs attention sharing, and similar to us, employs DIFT-aligned~\cite{tang2023emergent} feature maps for consistent image generation.
Several studies have explored the use of diffusion features for image interpolation, such as \citeN{samuel2024interpolation} that investigates various interpolation strategies in the noise space of diffusion models, or \citeN{he2024aid} that examines linear interpolation within the attention space to achieve realistic inbetweening.
In \papernospace, we apply correspondence to the inverted latents of an image to enable edits involving significant structural changes.
Also, interpolation is performed on \emph{aligned attentions} to balance content preservation and editing flexibility.

\subsubsection*{\textbf{Inversion in diffusion models}}
One way to edit images involves inverting the diffusion model: 
determining the noise that, when fed to the diffusion model, produces the source image~\cite{song2022denoisingdiffusionimplicitmodels}.
Some methods focus on improving this inversion process for more effective editing.
DDIM Inversion reverses the deterministic DDIM denoising process~\cite{song2022denoisingdiffusionimplicitmodels},
but accumulates small errors during inversion; these errors can lead to significant content drift, especially when using classifier-free guidance~\cite{ho2022classifierfreediffusionguidance} with high guidance~\cite{mokady2022null}.
To address this issue, some works rely on iteratively optimizing the intermediate noisy images~\cite{garibi2024renoise, AIDI, li2024source}.
Another line of inversion algorithms~\cite{HubermanSpiegelglas2023, cyclediffusion} predicts \textit{both} the initial as well as intermediate noise maps in the DDPM process, enabling better reconstruction and higher editing quality.
In our work, we build upon the DDPM inversion proposed by~\citeN{HubermanSpiegelglas2023} for its speed, high reconstruction quality, and versatility across editing tasks.

\subsubsection*{\textbf{Few step diffusion models}}
Because of the iterative nature of the diffusion process, these methods are usually slow, requiring 20-100 forward passes of the model.
Recent research have focused on designing few-step frameworks using distillation~\cite{yin2024onestep, salimans2022progressive}, consistency constraint~\cite{luo2023latent, xiao2023ccm, song2023consistency} and adversarial training~\cite{sauer2024fasthighresolutionimagesynthesis}, enabling image generation in 1-8 steps.
While these methods are faster, adapting them for editing is not trivial.
Several recent works attempted to adapt few-step diffusion to editing tasks.
\citeN{wu2024turboedit} trains an encoder for fast inversion, \citeN{xu2023infedit} proposes a novel inversion-free framework, and~\citeN{deutch2024turboedittextbasedimageediting} adapts editing-friendly DDPM inversion to few-step models.
These methods focus on \textit{appearance} changes with minimal structural edits, while the proposed \paper supports both.

%% file: sections/3_method.tex
\newcommand{\bwd}{\text{bw}}
\newcommand{\fwd}{\text{fw}}
\newcommand{\aln}{\text{aln}}
\section{Preliminaries}
\label{sec:preliminary}

\subsubsection*{\textbf{Noise-inversion}}
\citeN{HubermanSpiegelglas2023} maps an input image $x_0$ to 
\begin{align}
\{x_T, z_{T-1},\dots, z_1, z_0\},
\end{align}
where $x_T$ represents the inverted noise at the final timestep $T$, and $z_t$ denotes correction terms.
Given $x_0$, the algorithm first computes its noisy versions across all timesteps in \textit{forward} (fw) diffusion: 
\begin{equation} x_t^\fwd = \sqrt{\bar{\alpha}_t} x_0 + \sqrt{1 - \bar{\alpha}_t} \tilde{\epsilon}_t, \quad t = 1, \dots, T, \end{equation}
where $\tilde{\epsilon}_t$ is independent standard Gaussian noise, and $\bar{\alpha}_t$ is a parameter of the diffusion scheduler.
In the \textit{backward} (bw) process, the scheduler calculates $x_{t-1}^\bwd$ as:
\begin{equation}
x_{t-1}^\bwd = \mu_t(x_t^\bwd, c) + \sigma_t z_t,
\end{equation}
where $c$ is the text prompt used for inversion, and
\begin{equation}
\mu_t(x_t^\bwd, c) = \frac{1}{\sqrt{\alpha_t}} \left( x_t^\bwd - \frac{1 - \alpha_t}{\sqrt{1 - \bar{\alpha}t}} \epsilon_\theta(x_t^\bwd, t, c) \right)
\end{equation} 
is the predicted $x_{t-1}$ given $x_t^\bwd$, $\epsilon_\theta(x_t^\bwd, t, c)$ is the output of the diffusion model, and $\sigma_t$ is the variance of the scheduler.

At first glance, it is evident that if $z_t$ is standard Gaussian noise, as in the typical denoising process, then $x_{t-1}^\bwd$ and $x_{t-1}^\fwd$ will not necessarily match.
However, by setting $z_t$ as:
\begin{equation} 
z_t = \frac{x_{t-1}^\fwd - \mu_t(x_t^\bwd, c)}{\sigma_t}, 
\label{eq:friendly}
\end{equation}
we ensure $x_{t-1}^\bwd \equiv x_{t-1}^\fwd$.
Therefore, using the same text prompt~$c$ for generation leads to perfect reconstruction.
Modifying the text prompt to a desired edit $\hat{c}$ enables editing, provided the edit does not involve significant structural changes.

\subsubsection*{\textbf{TurboEdit}}

\citeN{deutch2024turboedittextbasedimageediting} introduce a series of modifications to the inversion algorithm to adapt it for the \textit{few-step} setting, including time-shifted inversion, norm-clipping at the final denoising step, and text-guidance to improve prompt alignment. Our method incorporates these modifications.

\section{Method}
\label{sec:method}
We build \paper on top of the few-step diffusion model (i.e., SDXL-Turbo~\cite{sauer2024fasthighresolutionimagesynthesis}), enabling both appearance editing and structural changes.

\subsubsection*{\textbf{Outline}}
In Section~\ref{sec:correspondence} we observe that using noise inversion for structural editing introduces artifacts~(Fig.~\ref{fig:dog1}).
We explain how and why a hierarchical correspondence-aware latent correction can resolve these issues.
Second, edits that involve significant structural changes or object additions require generating entirely new content or revealing regions absent in the source image.
Directly using the keys and values of the source image in the self-attention modules of the diffusion model, as done in MasaCtrl~\cite{Cao_2023_ICCV}, often results in unwanted artifacts or misalignment between the target image and appearance editing instructions in the text-prompt~(Fig.~\ref{fig:att_strategy}:b).
To mitigate this, it is necessary to combine the keys and values of both the source and target images. In Section~\ref{sec:spherical}, we explore various strategies for this combination and introduce a novel method called \textit{correspondence-aware attention interpolation}.
We propose two strategies for interpolation: spherical (SLERP) and linear, and show that SLERP is beneficial.
Finally, in Section~\ref{sec:structure}, we demonstrate that by matching the queries of the source and target images, we enable control over the extent of structural change in the target image.

\subsection{Correspondence-aware latent correction}
\label{sec:correspondence}
As explained in Section~\ref{sec:preliminary}, noise-inversion maps the input image $x_0$ to~$\{x_T, z_{T-1}, \dots, z_1, z_{0}\}$, where $\{z_t\}$ serve as corrections, ensuring perfect reconstruction when the text prompt $c$ used for inversion \textit{matches} the one used for generation.
However, when the text prompt requires a large deformation of the source image, the corrections~$\{z_t\}$ are not pixel-aligned with respect to the generated image, leading to severe artifacts~(Fig.~\ref{fig:dog1}).
%
%
To address this issue, we align the corrections~$\{z_t\}$ with the spatial transformation introduced in the edited image in the final two steps of denoising.
This involves constructing a correspondence map $C_{T\rightarrow S}$ between the source image $I_S$ and the target image $I_T$.
We achieve this via their DIFT features~\cite{tang2023emergent}, denoted as $D_S$ and $D_T$:
\begin{equation}
C_{T\rightarrow S}(p) = \arg\max_{q \in I_S} 
\:\: \text{sim}(D_T(p), D_S(q)),
\end{equation}
where $p$ and $q$ are pixels from the target and source images, respectively, and $\text{sim}$ is the cosine similarity between the two feature vectors.
We construct an aligned correction term $z_t^\aln$ as:
\begin{equation}
z_t^\aln(p) = z_t(C_{T\rightarrow S}(p)).
\end{equation}

\subsubsection*{\textbf{Patch correspondence}}
Since DIFT features might be noisy and inaccurate, the resulting target image often contains artifacts({inset - left). Therefore, we propose a patch-wise correspondence approach.
The DIFT features $D_S$ and $D_T$ are divided into small, overlapping patches, and correspondence is computed for each patch rather than individual pixels.
For each patch, we concatenate its pixel-wise DIFT features and calculate cosine similarity between these concatenated features.
Due to overlapping patches, multiple patches may contribute to the alignment of a single pixel $p$.
To obtain the final aligned correction latent $z_t^\aln(p)$, we average the contributions from all overlapping patches at pixel $p$.

\setlength{\intextsep}{.0em}
\setlength{\columnsep}{1.5em}
\begin{wrapfigure}[8]{r}{0.6\linewidth}
    \centering
    \includegraphics[width=\linewidth]{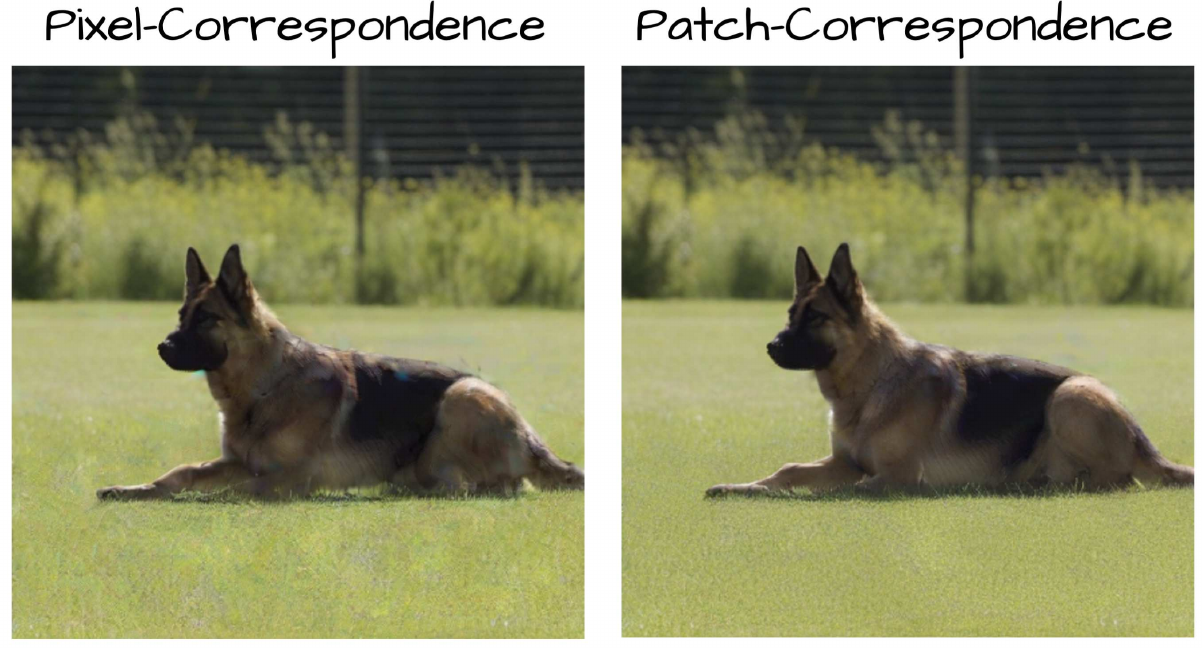}  \label{fig:inset}
\end{wrapfigure}

As denoising progresses and the features become less noisy at later timesteps, the size of the patches is gradually reduced.
This ensures that the correspondence becomes more precise, adapting dynamically to the evolving reliability of the features~(see inset -right).

\subsection{Correspondence-aware attention interpolation}
\label{sec:spherical}
Achieving high-quality image editing requires balancing the preservation of key aspects of the source image (e.g., appearance or identity) with the introduction of new elements or modifications. 
Recent approaches often achieve this by injecting the attention features of the source image into the denoising process of the target image~\cite{Cao_2023_ICCV}.
While this method is effective, it overlooks the fact that editing often involves generating \textit{new} content, and this content may lack clear correspondence to content stored within the source image.
We now consider existing methods, and present a novel strategy for combining attentions between source and target images.

\subsubsection*{\textbf{Mutual self-attention}}
MasaCtrl~\cite{Cao_2023_ICCV} uses the source image's keys and values in the self-attention modules of the diffusion model.
This ensures that the context of the source image,  such as its appearance and identity are preserved: 
\begin{equation}
f^e = \text{Attention}(Q_T, K_S, V_S),
\end{equation} 
where $Q_T$ is the query of the target image, $K_S$ and $V_S$ are the keys and values of the source image, and $f^e$ represents the output of the self-attention module.
While this strategy effectively retains the appearance and identity of the original content, it also limits the model's ability to generate \textit{new} content, such as adding objects or significantly altering appearances~(see Fig.~\ref{fig:att_strategy}:b).

\subsubsection*{\textbf{Concatenation}}
An approach to incorporate appearance editing is concatenating the keys and values of the source and target images as done by ~\citeN{hertz2023StyleAligned}, ~\citeN{deng2023zstar}, and~\citeN{tewel2024consistory}: 
\begin{equation} 
f^e = \text{Attention}\bigl(Q_T, [\lambda \cdot K_S, K_T], [\lambda \cdot V_S, V_T]\bigr), 
\end{equation} 
where $[~,~]$ denotes concatenation, and $\lambda$ is a scaling factor that controls how much the source appearance affects the target image.
While this enables appearance changes, it often fails to achieve a smooth interpolation between the two appearances.
This can result in unnatural ``appearance leakage'', such as elements of the \textit{red car} blending into the \textit{white bus}~(see Fig.~\ref{fig:att_strategy}:c).

\newcommand{\lerp}{\mathcal{L}}
\newcommand{\slerp}{\mathcal{Q}}
\newcommand{\magint}{\mathcal{M}}
\newcommand{\magslerp}{\magint\slerp}

\subsubsection*{\textbf{Linear interpolation}}
Another approach is to linearly interpolate the keys and values of the source and target images~\cite{he2024aid}.
\setlength{\intextsep}{.0em}
\setlength{\columnsep}{1.5em}
\begin{wrapfigure}[6]{r}{0.6\linewidth} 
    \vspace{-1pt}
    \centering
     \includegraphics[width=\linewidth]{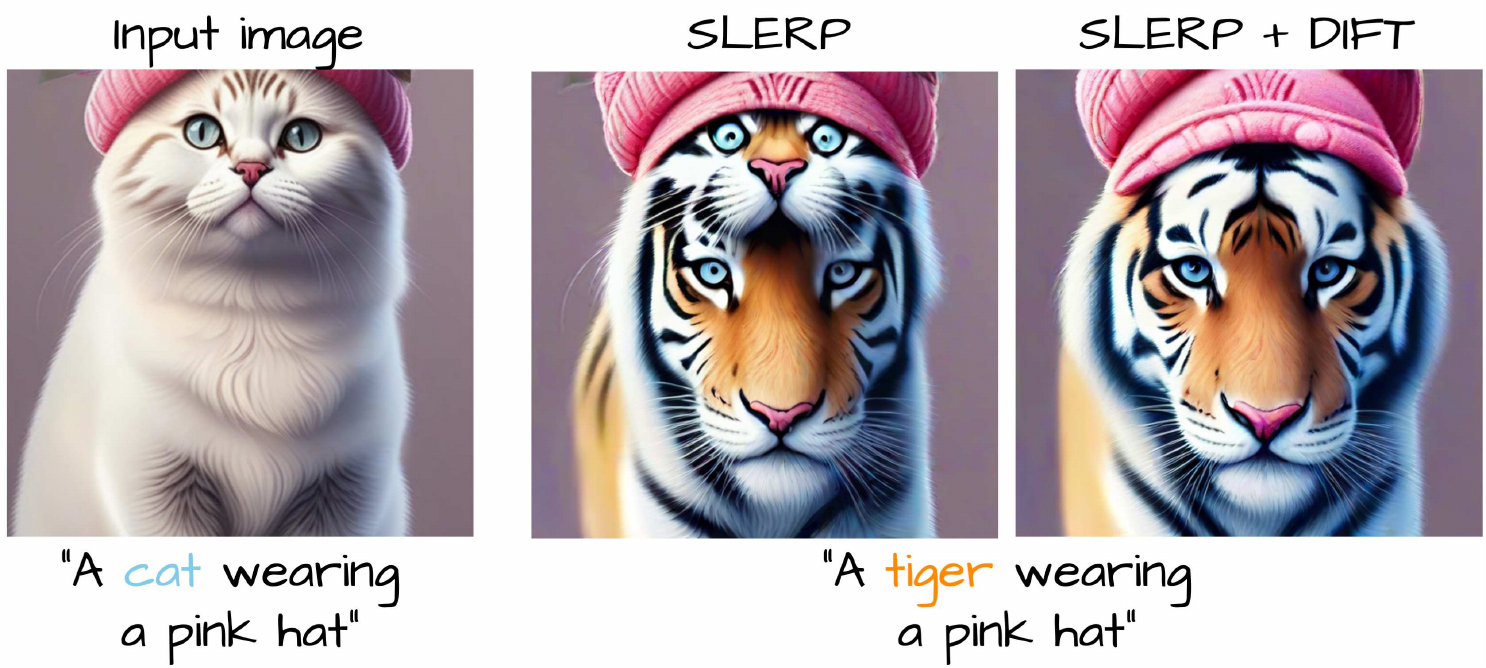} 
    \label{fig:inset}
\end{wrapfigure}
Differently from \citeN{he2024aid}, we linearly interpolate after matching features between source and target features (Sec.~\ref{sec:correspondence}), as otherwise this may cause artifacts due to the mis-alignment of source and target (see the inset):
\begin{gather} 
\lerp(v_1, v_2, \alpha) = (1-\alpha) \cdot v_1 + \alpha \cdot v_2,
\\
f^e = \text{Attention}(Q_T, \lerp(K_S^{\text{aln}}, K_T, \alpha), \lerp(V_S^{\text{aln}}, V_T, \alpha)).
\end{gather}

\begin{figure} [t]
  \includegraphics[width=0.99\linewidth]{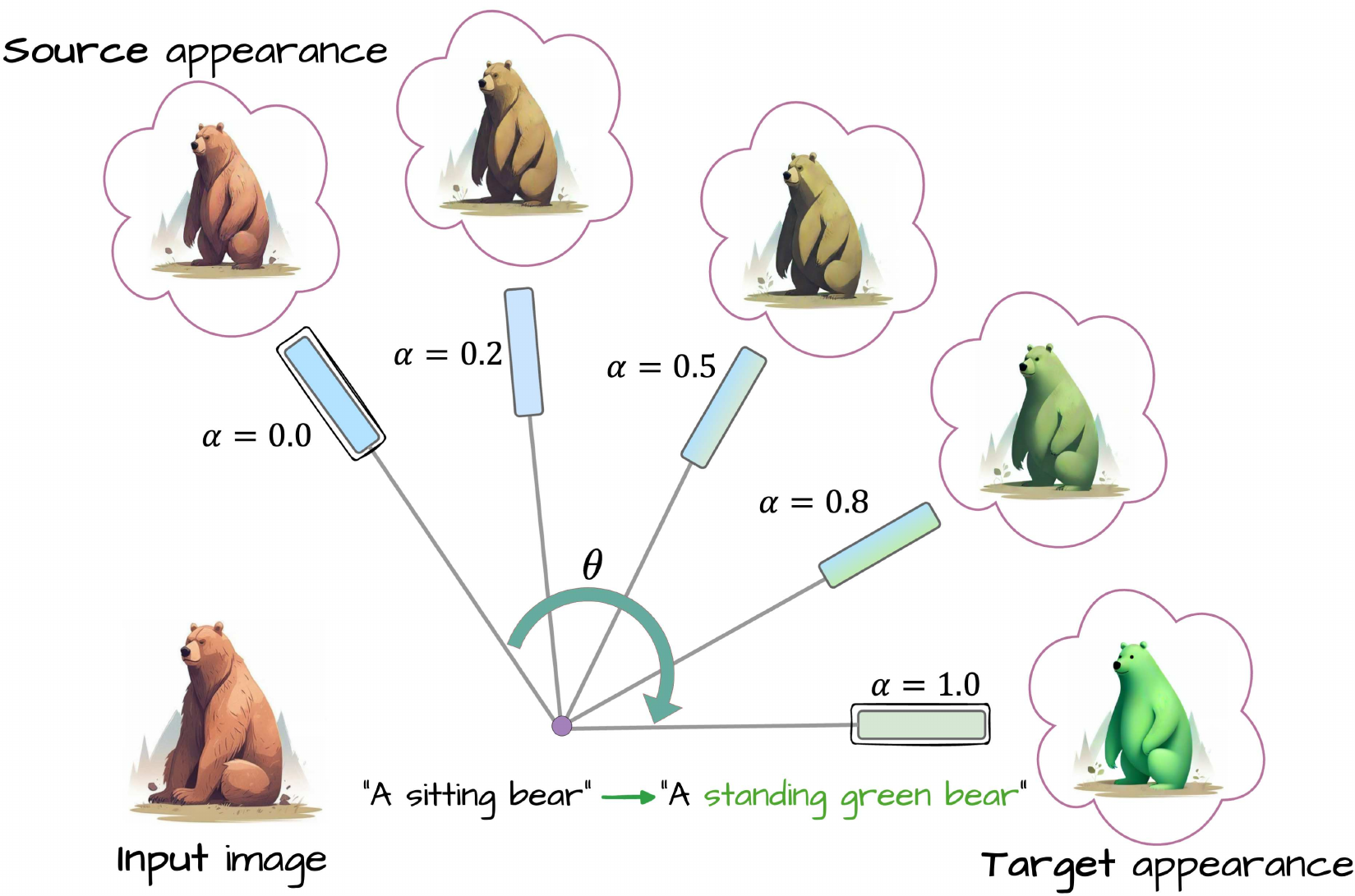}
  \caption{\textbf{Adjusting $\alpha$} Provides control over the appearance transition between the source image and the target appearance. When $\alpha = 0$, the appearance is entirely derived from the source image, and when $\alpha = 1$, it fully reflects the editing prompt. Intermediate values of $\alpha$ allow for a gradual blend, enabling fine-grained control between these two extremes.
  }
  \Description{Teaser Figure.}
  \label{fig:alpha}
\end{figure}
\noindent
While this approach is somewhat effective, it sometimes causes unwanted artifacts when interpolating between features that are significantly different (see the mirrors in Figure~\ref{fig:att_strategy}:d).
To address this limitation, we explore the use of \textit{spherical linear interpolation (SLERP)} for interpolating between the keys and values, which takes vector directions into account for smoother blending:
\begin{equation} 
\slerp(v_1, v_2; \alpha) = \tfrac{\sin((1 - \alpha) \theta)}{\sin(\theta)} \cdot \frac{v_1}{|v_1|} + \tfrac{\sin(\alpha \theta)}{\sin(\theta)} \cdot \frac{v_2}{|v_2|},
\end{equation}
where $\theta$ is the angle between the vectors $v_1$ and $v_2$, and $\alpha \in [0,1]$ is the blending weight. The output of this operation is a unit vector. To preserve the magnitude information of the vectors, we also interpolate their magnitudes as $\magint(v_1, v_2; \alpha) = (1 - \alpha)\cdot|v_1| + \alpha \cdot |v_2|$.
Combining these, the full interpolation formula becomes: \begin{equation} \magslerp(v_1, v_2; \alpha) = \magint(v_1, v_2; \alpha) \cdot \slerp(v_1, v_2; \alpha). \end{equation} In the self-attention modules, this results in: 
\begin{equation}
f^e = \mathrm{Attention}\Bigl( 
    Q_t, \; \magslerp(K_S^{\text{aln}}, K_T; \alpha),
    \;    \magslerp(V_S^{\text{aln}}, V_T; \alpha) 
\Bigr).
\end{equation}
This ensures that transitions between source and target vectors respect their angular relationships and provides smoother and more reliable blending between the appearance of the source and target images (see Fig.~\ref{fig:att_strategy}:e). Furthermore, adjusting $\alpha$ enables \textit{control} over the extent of appearance changes in the target image relative to the source image (see Fig.~\ref{fig:alpha}).

\begin{figure}[t]
    \centering
    \includegraphics[width=\linewidth]{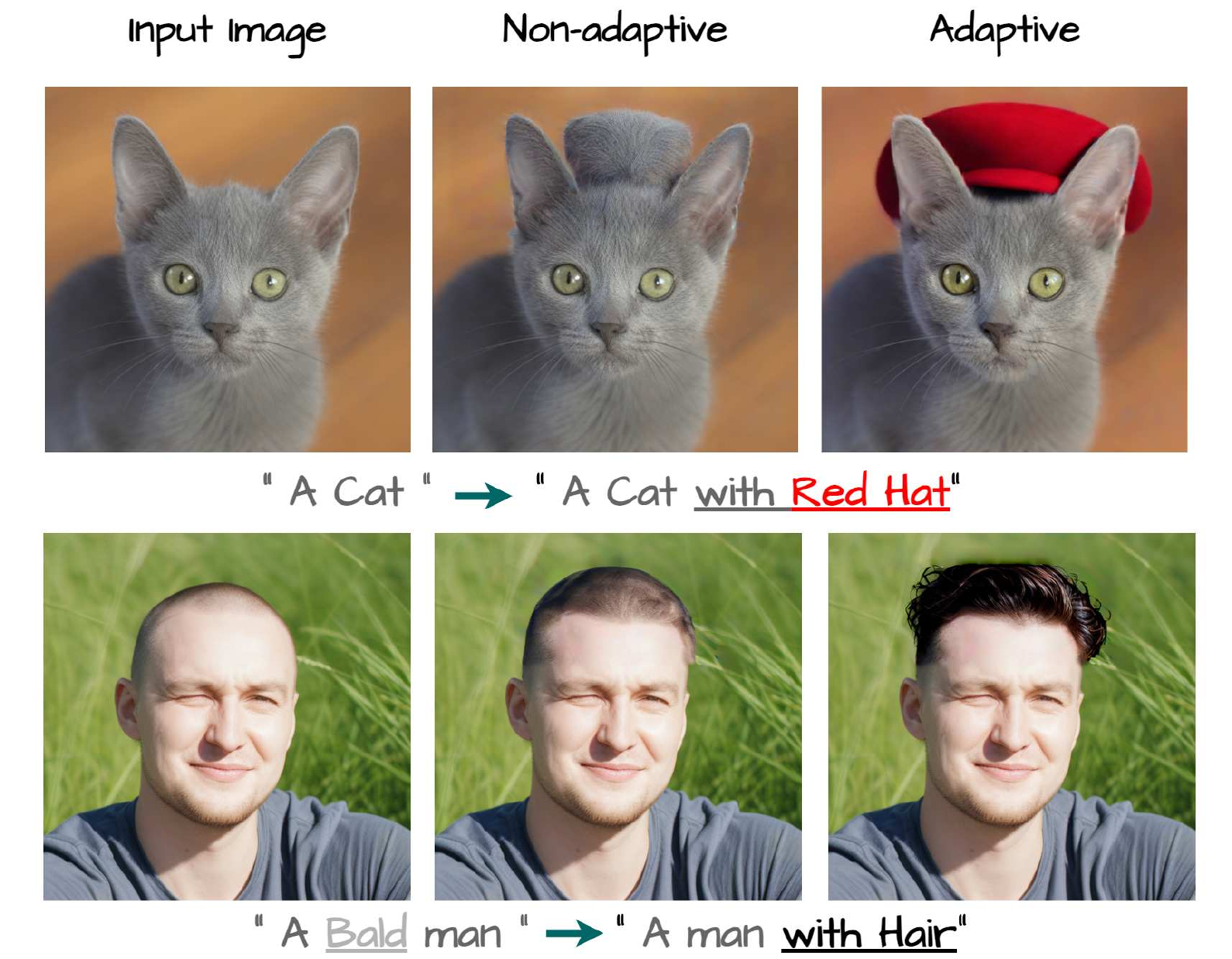}\vspace{-3pt}
    \caption{\textbf{Effect of content-adaptive interpolation.} Interpolating the keys and values for target image regions without clear correspondence in the source image results in undesirable edits (\textbf{b}). Classifying these regions and using only the target keys and values mitigates this issue (\textbf{c}).
    }
    \label{fig:discard}\vspace{-3pt}
\end{figure}

\subsubsection*{\textbf{Content-adaptive interpolation}}
In situations where the prompt suggests the insertion of a new object or a deformation that causes significant disocclusion, expecting that every generated/target pixel matches a pixel in the source image is not reasonable.
In such scenarios, naively aligning the keys and values of the source and target images leads to text misalignment and/or visual artifacts~(see Fig.~\ref{fig:discard}).
To address this issue, we propose to \textit{classify} pixels in the target image that do not have any correspondence in the source image (i.e.~``new'' pixels).
We achieve this by a \emph{bidirectional} comparison between source and target image patches.
Specifically, for each patch in the source image $s \in \mathcal{S}$, we compute the set $K(s)$ of k-nearest target patches:
\begin{equation}
    K(s) = \underset{t \in \mathcal{T}}{\mathrm{arg\,top}_k} \:\: \text{sim}(s, t),
\end{equation}
where $\text{sim}(s, t)$ is the cosine similarity from Section~\ref{sec:correspondence}.
Similarly, for each patch in the target image $t \in \mathcal{T}$, we define its top-$k$ nearest patches as $K(t)$.
We call a pair of patches $(s, t)$ \textit{bidirectionally matched} if~$t \in K(s)$ and~$s \in K(t)$.
For bidirectionally \textit{matched} patches, we use the hyper-parameter $\alpha$ specified by the user, while by setting $\alpha=1$ we could let patches being purely driven by the text, rather than the source image.
In particular, for \textit{unmatched} target patches $t \in \mathcal{U}$ we set $\alpha=1$ if we deem the correspondence to be particularly ``weak''.
We measure this weakness by computing the score of the best available match:
\begin{equation}
\text{sim}_{max}(t) = \max_{s \in \mathcal{S}} \:\: \text{sim}(s,t)
\end{equation}
and determining which subset of $t \in \mathcal{U}$ have the worst matches by computing the $\gamma$-quantile of the scores in this set:
\begin{equation}
\tau_\gamma = \text{quantile}_\gamma \bigl( \{ \text{sim}_{\max}(t) \mid t \in \mathcal{U} \} \bigr).
\end{equation}
finally setting $\alpha{=}1$ whenever $sim_{max}(t) {<} \tau_\gamma$.
We typically set $\gamma {=} 3\%$ in our experiments.

\begin{figure}[t]
    \centering
    \includegraphics[width=\linewidth]{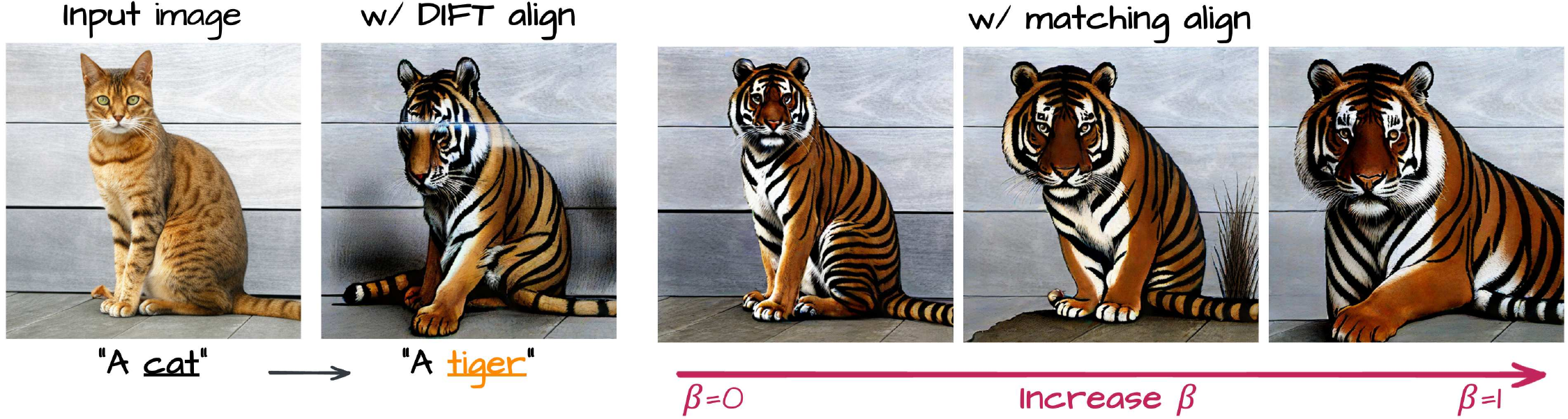}\vspace{-3pt}
    \caption{\textbf{Structure Alignment.} DIFT-aligned queries produce unnatural edits, while our matching algorithm with adjustable blending weights ($\beta$) enables transitions between full structure alignment and new layout generation.
    }\label{fig:align}\vspace{-3pt}
\end{figure}

\subsection{Structural alignment}
\label{sec:structure}
Preserving the overall layout of the image (i.e. preserving structure) is important when editing images.
Recent works~\cite{Cao_2023_ICCV, alaluf2023crossimage} have demonstrated that it is the queries in the self-attention modules of diffusion models that specify the structure of the generated image.
Hence, while in Section~\ref{sec:spherical} we described key/value mixing, we now incorporate \textit{queries} from the source image during the denoising process to retain the overall image structure.
Our key idea is that reproducing the structure of the original image, up to a non-rigid deformation, implies that we want to find \textit{all} the local structures of the source image within the generated target.
We achieve this via Hungarian matching~\cite{hungarian} between source and target queries, as this provides us with a \textit{one-to-one} matching~(i.e.~every target query should match one source query).
In particular, Hungarian matching computes the optimal permutation given a weight matrix $C$, which then shuffles our generation queries:
\begin{align}
    \pi^* = \argmin_{\pi \in \text{Perm}(N)} \:\: \sum_{n=1}^N C[n, \pi(n)], \quad
    Q_T^\pi[n] \leftarrow Q_T[\pi^*(n)]  
\end{align}
We define our weight matrix $C$ as a linear interpolation controlled by a blending weight $\beta$ between two matrices (described next):
\begin{equation}
    C = (1 - \beta)\cdot \text{normalize}(C_{\text{SA}}) + \beta \cdot \text{normalize}(C_{\text{TC}}),
\end{equation}
where $\text{normalize}(.)$ rescales the matrices to ensure comparability.
The first matrix $C_{\text{SA}}$ encourages the target queries to remain similar to the source queries, hence promoting \textit{Source Alignment}:
\begin{align}
    C_{\text{SA}}[i, j] \;=\; 1 \;-\; \text{sim}(q_s[i] , q_t[j]).
\end{align}
The second matrix $ C_{\text{TC}}$ attempts to penalize index differences among the target queries, hence promoting \textit{Target Consistency}:
\begin{align}
    C_{\text{TC}}[i, j] = \sqrt{|\,i - j\,|}.
\end{align}
Varying \( \beta \) provides us with control over the structure of the target image, transitioning between preserving the source structure ($\beta{\approx}0$) or adhering more to the text prompt while remaining self-consistent ($\beta{\approx}1$).
An illustration of the effect of $\beta$ can be found in Figure~\ref{fig:align}.
Note that this process is limited to the first step of denoising when the coarse structure of the generated image is established.

\begin{figure}[t]
  \includegraphics[width=\linewidth]{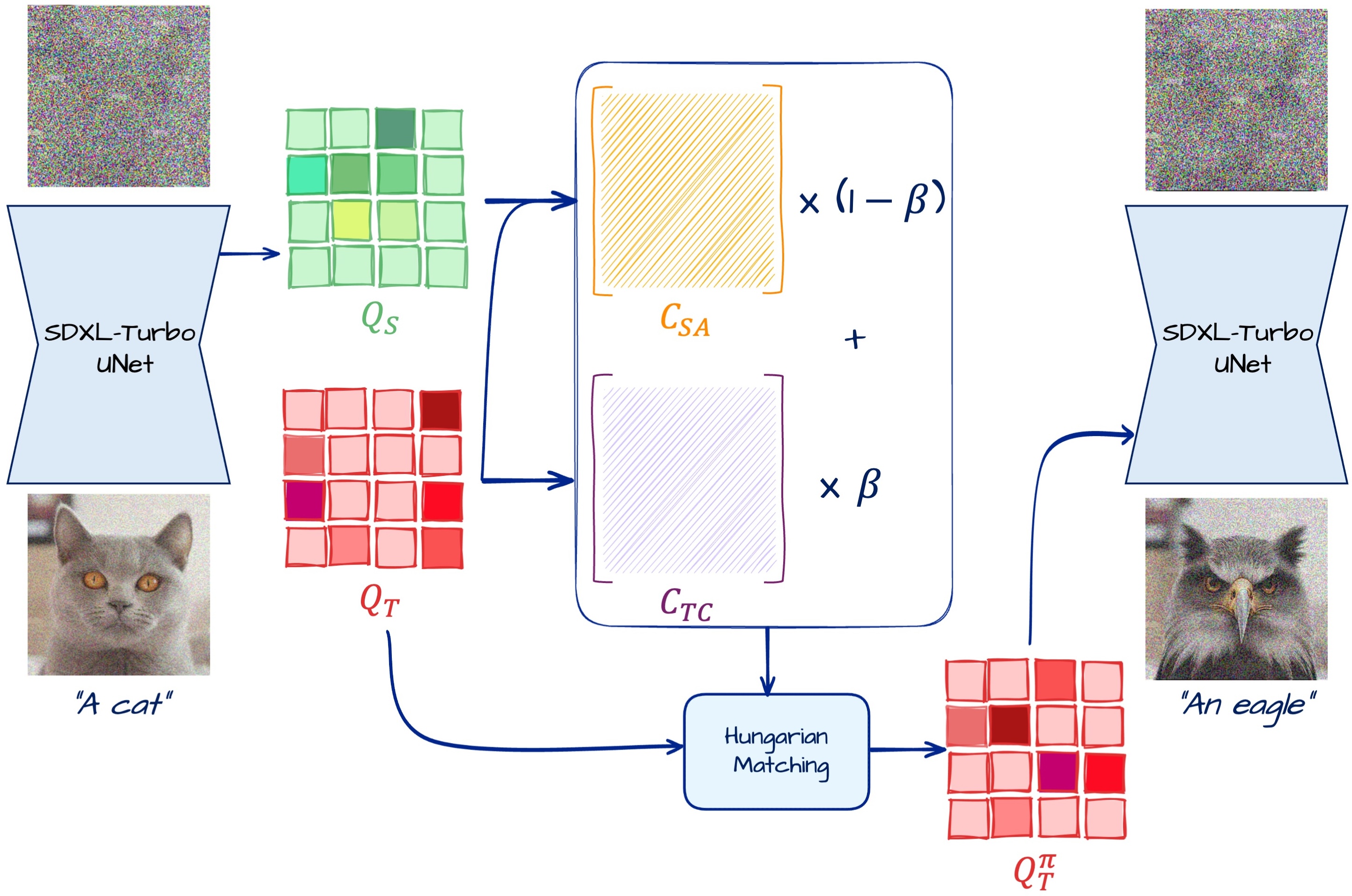}
  \caption{\textbf{Structural Alignment.} In the first denoising step, we extract self-attention queries from both source and target images.
We then define two cost matrices: \(C_{SA}\), which promotes structural alignment between source and target, 
and \(C_{TC}\), which preserves target structure. By linearly combining these matrices, we can control the strength of alignment. The resulting cost matrix is then used in the Hungarian matching algorithm to permute the target queries, aligning them with the source's structure.
}
  \Description{Teaser Figure.}
  \label{fig:hungarian}
\end{figure}

%% file: sections/4_experiments.tex
\section{Experiments}
\label{experiments}

\begin{figure*}
  \includegraphics[width=0.83\textwidth]{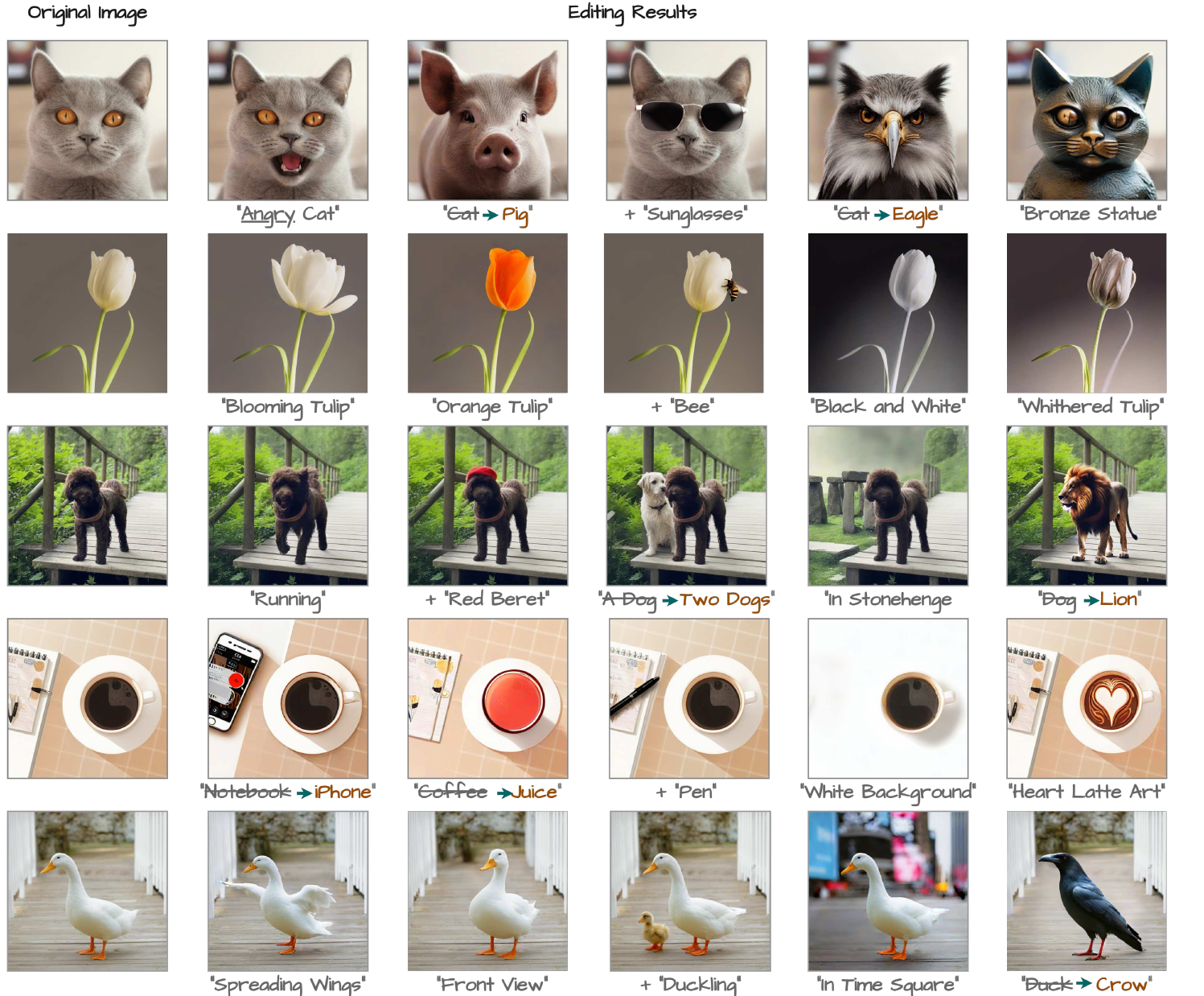}
  \caption{\textbf{Qualitative results}. We demonstrate the ability of our method to perform various types of edits on multiple images.}
  \label{fig:qual}
\end{figure*}

In this section, we present various editing results on real images, demonstrating the versatility of our method. We then compare our approach with both multi-step and few-step baselines to highlight its advantages in terms of visual quality and speed. Finally, we conduct ablation studies to analyze the contribution of each component.

\noindent\textbf{Qualitative Results.}
Fig.~\ref{fig:qual} showcases several edits generated by our 4-step denoising procedure. These examples include non-rigid deformations, inserting new objects, and replacing existing objects. Our method generally preserves the overall structure of the input image while accurately reflecting the requested edits.

To compare with existing approaches, Figure~\ref{fig:comparison} illustrates that our approach is more successful at maintaining the subject’s identity and reducing artifacts. We focus on few-step baselines such as TurboEdit~\cite{wu2024turboedit} and InfEdit~\cite{xu2023infedit} as they operate in a similar few-step regime, as well as multi-step frameworks such as MasaCtrl~\cite{Cao_2023_ICCV} and Edit friendly DDPM inversion~\cite{HubermanSpiegelglas2023}. Our results exhibit fewer distortions and better fidelity, particularly upon closer inspection.

We further expand the comparison to multi-step methods, including Prompt-to-Prompt (P2P)~\cite{hertz2022prompt}, plug-and-play (PnP)~\cite{tumanyan2022plugandplay}, instruct-pix2pix~\cite{brooks2023instructpix2pix}, and StyleDiffusion~\cite{li2023stylediffusion} (see Figure~\ref{fig:comparison_extra}). Despite being significantly faster, our few-step approach achieves comparable or superior results in preserving details and adhering to the edits.

\input{tables/user_study_comp}
\input{tables/user_study_abl}

\noindent\textbf{User studies.}
Although we quantitatively show in Supplementary that across different metrics \paper is comparable or superior to alternatives, standard metrics (e.g., PSNR and LPIPS) often fail to capture the visual quality of edits with significant deformation—the key focus of our paper. Therefore, we also conducted a user study. Participants were shown the original image, the editing prompt, and outputs from our method and various baselines. They ranked the images based on (i) alignment with the prompt and (ii) preservation of the subject in the source image, using a scale from 1 (worst) to 4 (best). The average rankings are summarized in Tab.\ref{tab:user_study_comp}. Feedback from 51 participants strongly favored our method over other few-step approaches and found it comparable to computationally intensive multi-step techniques. Additionally, a separate user study on attention mixing strategies revealed that correspondence-aligned SLERP interpolation produced the best results, as shown in Tab.\ref{tab:user_study_abl}.

\noindent\textbf{Ablation Studies.}
We now examine the contributions of the main components in our framework:

     \textit{Structure Alignment.} Disabling structure alignment reduces background fidelity,although the edited object remains well-aligned to the text prompt. Visual comparisons (see Supplementary) confirm that structure alignment is crucial for preserving scene details.
    
    \textit{Correspondence-Aware Latent Correction.} Removing this module causes significant distortions in the edited region. Hence, the latent correction is essential for producing coherent edits. For visual results, see Supplementary.
    
    \textit{SLERP vs. LERP.} While switching from SLERP to LERP often produces similar outcomes, SLERP can yield more consistent transitions in certain challenging cases. For visual results, see Supplementary.    
    \textit{Removing correspondence alignment From Attention.} As seen in Sec.~\ref{sec:spherical}, this leads to more artifacts, as it helps enforce alignment between the modified content and the background.

%% file: tables/user_study_comp.tex
\begin{table}[t]
	  \centering
    \captionsetup{font=small}
    \caption{\textbf{User study}. Our method has received a significantly higher score than the alternatives.}
    \small
	\begin{tabular}{l c}
        Method & Average Ranking ($\uparrow$)\\
		\midrule
        MasaCtrl~\cite{Cao_2023_ICCV}  & 1.02\\
        DDPM Inversion~\cite{HubermanSpiegelglas2023} & 1.78\\
        InfEdit~\cite{xu2023infedit} & 1.67\\
        TurboEdit~\cite{deutch2024turboedittextbasedimageediting} & 2.24\\
        Ours  & \textbf{3.29}\\
	\end{tabular}

    \label{tab:user_study_comp}
\end{table}

%% file: tables/user_study_abl.tex
\begin{table}[t]
	  \centering
    \captionsetup{font=small}
    \caption{\textbf{User study}. Ablation user study on different attention mixing strategies mentioned in Sec.~\ref{sec:spherical}. It is evident that correspondence-aligned SLERP interpolation yields the best results.}
    \small
	\begin{tabular}{l c}
        Method & Average Ranking ($\uparrow$)\\
		\midrule
        Mutual & 0.40\\
        Concatenation & 1.71\\
        LERP & 2.08\\
        LERP+DIFT & 2.58\\
        SLERP+DIFT  & \textbf{3.23}\\
	\end{tabular}
    \label{tab:user_study_abl}
\end{table}

%% file: sections/5_conclusion.tex
\section{Conclusion, Limitations, Future Work}
\label{conclusion}
We have introduced \papernospace, a novel diffusion-based image editing method that addresses the challenges of structural edits, such as non-rigid transformations and object insertions. By leveraging innovative attention mixing and correspondence-aware techniques, our approach enables accurate texture preservation and structural alignment. Unlike existing methods, \paper effectively generates new content when required while maintaining fidelity to the source image where relevant. 
Our results demonstrate significant improvements in visual quality and flexibility across a wide range of editing tasks, including pose alterations, object manipulation, and texture modifications. Quantitative evaluations further validate that \paper consistently outperforms state-of-the-art methods in both quality and versatility. However, our method still suffers from some limitations. For example, prompts may change unintended parts of the image (e.g., changing a car's color might also affect the background). This can be resolved by using masks automatically obtained via cross and self-attention. While this is a promising direction, it is challenging with only four steps denoising and can be considered as a future work. Another future directions could be to extend \paper to video editing or evaluate alternative non-linear interpolation techniques for attentions.




%% file: sections/supplementary.tex
\input{tables/quant_comp2}

\section{Implementation details}  
\label{sec:hf_supp}  

We use SDXL-Turbo~\cite{sauer2024fasthighresolutionimagesynthesis} from the Diffusers library~\cite{diffusers} as our few-step diffusion model. Both the inversion of the source image and the generation of the target image are performed using \textit{four} denoising steps. Following FlexiEdit~\cite{koo2024flexiedit}, we observed that high-frequency components in the latent space during the initial timesteps of denoising can hinder flexible editing by imposing rigid constraints, limiting significant deviations from the original image’s structure. To facilitate larger and more flexible changes, we apply high-frequency suppression to the latent representation at the first timestep. Furthermore, while our method works without masking, incorporating a mask to specify regions of the image that should remain unchanged can enhance background preservation during editing. Following~\citeN{avrahami2020blendeddiffusion}, we use masked latent blending to preserve these regions by blending them from the source image. 

We applied our latent correction in timesteps 3 and 4 of denoising in which the patch sizes are $5\times5$ and $3\times3$ respectively.  Moreover, in our content-adaptive interpolation, we found that setting $k=3$ or $k=4$ typically produces better results.

\section{Quantitative results}
We perform a quantitative comparison of \paper with several few-step and multi-step editing methods. for content preservation, we measure PSNR, LPIPS~\cite{zhang2018lpips}, mean squared error (MSE), and structural similarity~\cite{wang2004ssim} losses between the backgrounds of the source and target images. For text alignment we measure CLIP similarity~\cite{radford2021clipsim} between the target image with and without subject masking and the editing text-prompt. The results are presented in Tab.~\ref{tab:quant2}.

\section{Additional Experiments}

\subsection{Ablation on Structure alignment}
\begin{figure*}[h]
    \centering
    \includegraphics[width=0.55\textwidth]{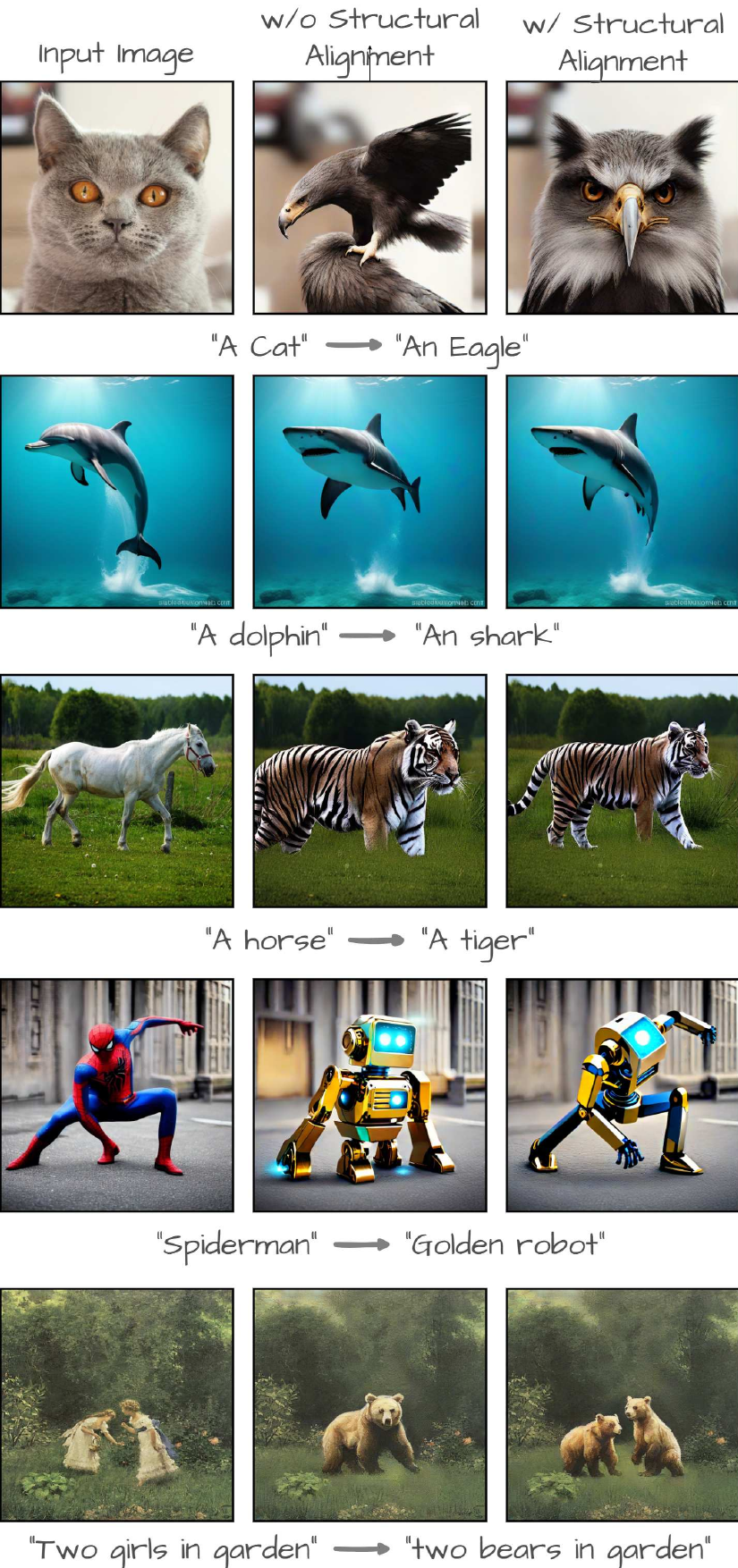}
    \caption{\textbf{Ablation on structure alignment.} By applying our structure alignment, we can preserve the structural layout of the source image.}
    \label{fig:supp_structure}
\end{figure*}

In Fig.~\ref{fig:supp_structure}, we show five examples to illustrate how our structural alignment step preserves the original pose and layout.
Each row compares results generated \emph{without} alignment (left) to those generated \emph{with} alignment (right).
Without alignment, the target images tend to deviate significantly from the source structure.
In contrast, applying our one-to-one query matching ensures that the coarse spatial arrangement of the source image is retained, resulting in edited images that faithfully reproduce the same pose while incorporating the desired changes.

\subsection{Ablation on latent correction}

\begin{figure*}[h]
    \centering
    \includegraphics[width=0.65\textwidth]{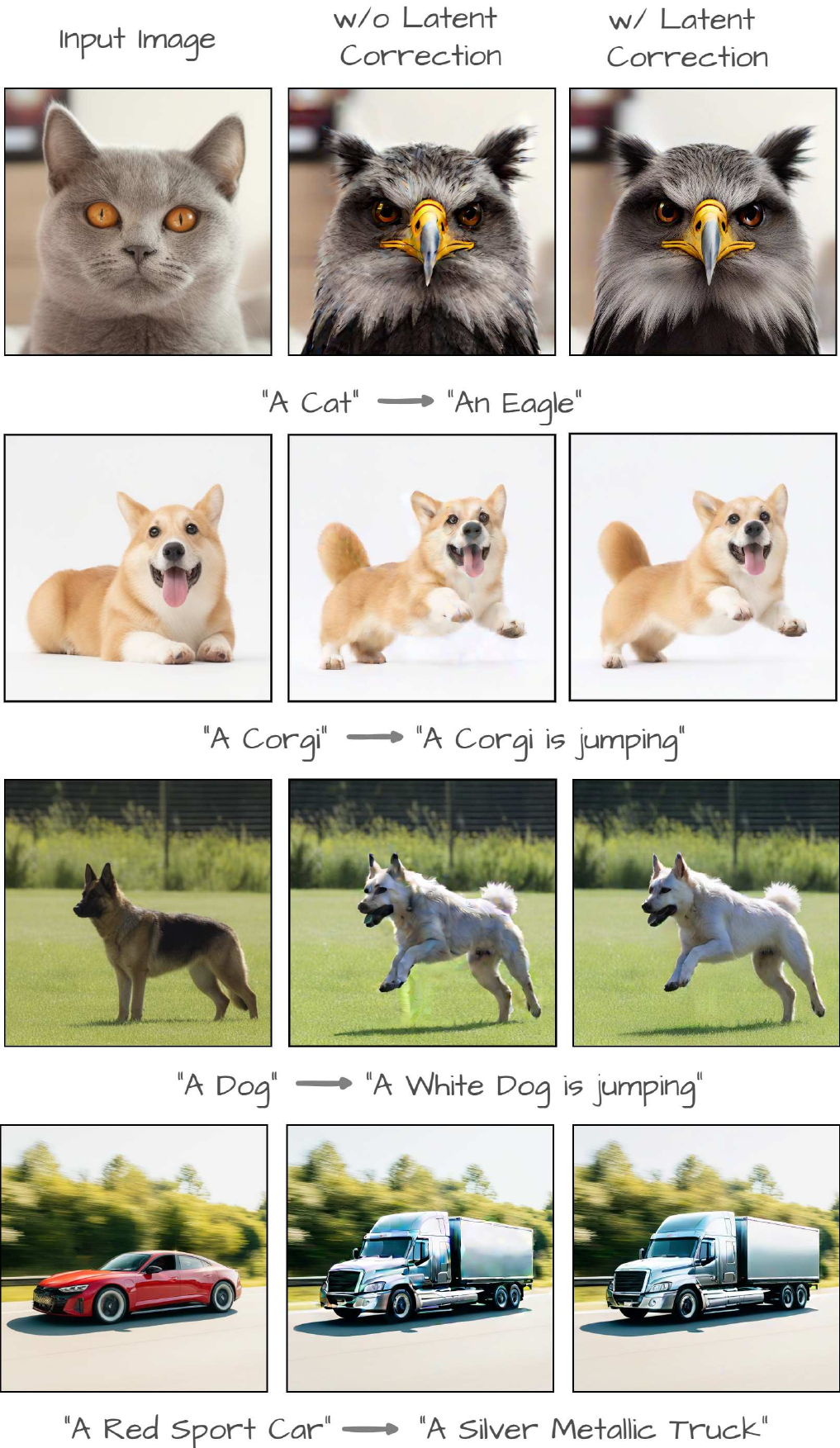}
    \caption{\textbf{Ablation on latent correction.} Without latent correction, multiple misalignment artifacts and unnatural deformations occur. Applying correction produces cleaner and more realistic results.}
    \label{fig:supp_correction}
\end{figure*}

 In Fig.~\ref{fig:supp_correction}, we compare our editing results \emph{with} and \emph{without} applying latent correction. When the correction is skipped, we frequently observe misalignment artifacts and unnatural deformations, especially for edits involving large shape changes. In contrast, incorporating our proposed patchwise correspondence alignment effectively mitigates these artifacts, leading to higher-quality edits that better preserve both the overall structure and fine details of the source image. This highlights the importance of the latent correction mechanism in achieving coherent and visually appealing results.

\subsection{Ablation on different attention mixing strategies}
\begin{figure*}[h]
    \centering
    \includegraphics[width=0.9\textwidth]{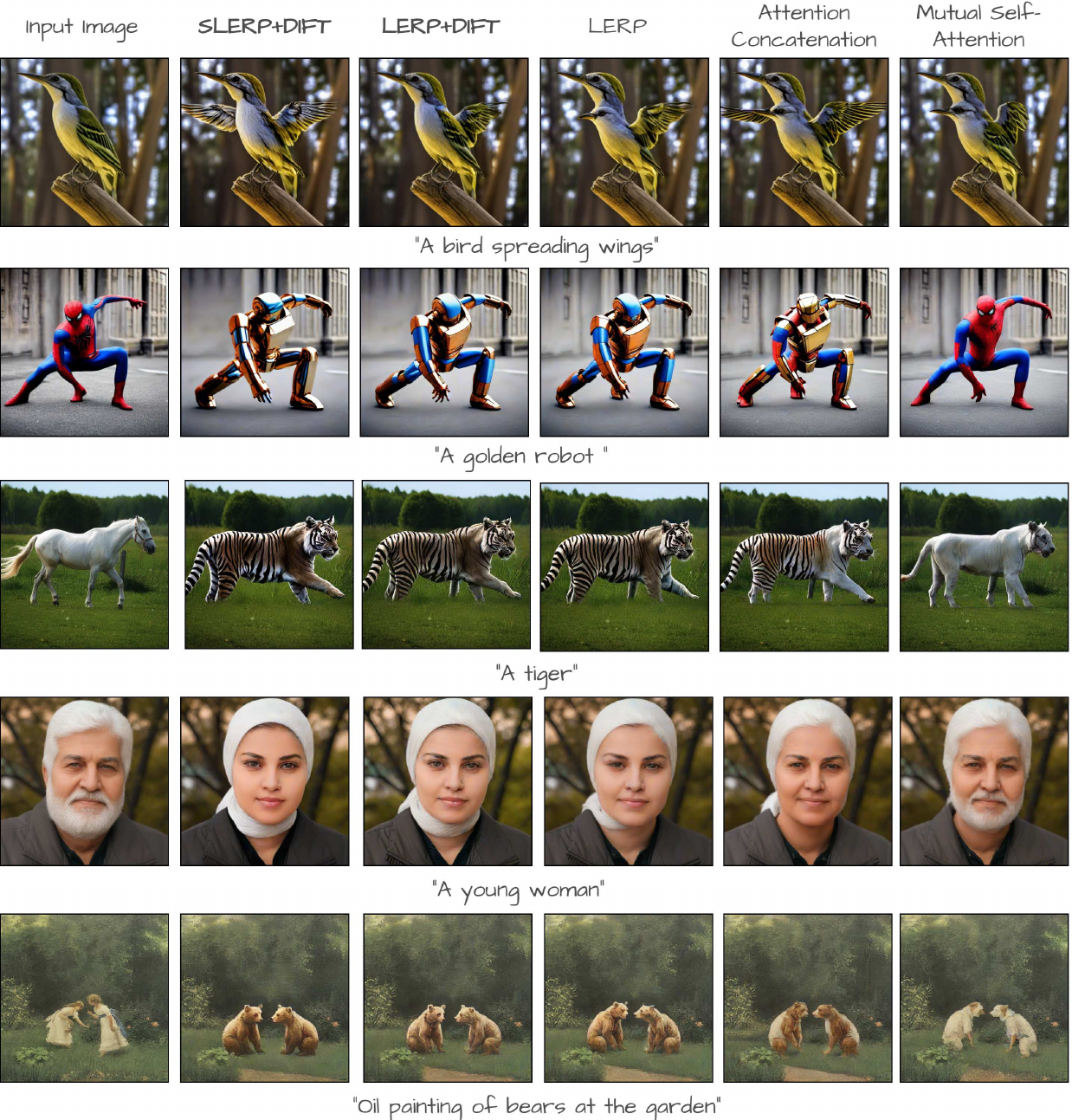}
    \caption{\textbf{Ablation on attention mixing strategies.} With these visual results, we demonstrate that DIFT-aligned SLERP yields the best results.}
    \label{fig:supp_attention}
\end{figure*}

In Fig.~\ref{fig:supp_attention}, we provide visual comparison between different attention mixing strategies. These include mutual self-attention~\cite{Cao_2023_ICCV}, Concatenating source and target image attentions~\cite{tewel2024consistory, hertz2023StyleAligned}, and our linear and spherical interpolation with and without DIFT alignment. As evident, the aligned spherical interpolation yields the most realistic and natural results.

\section{Generalization through Controllable Parameters}
\label{sec:generalization}
\begin{figure*}[h]
    \centering
    \includegraphics[width=0.7\textwidth]{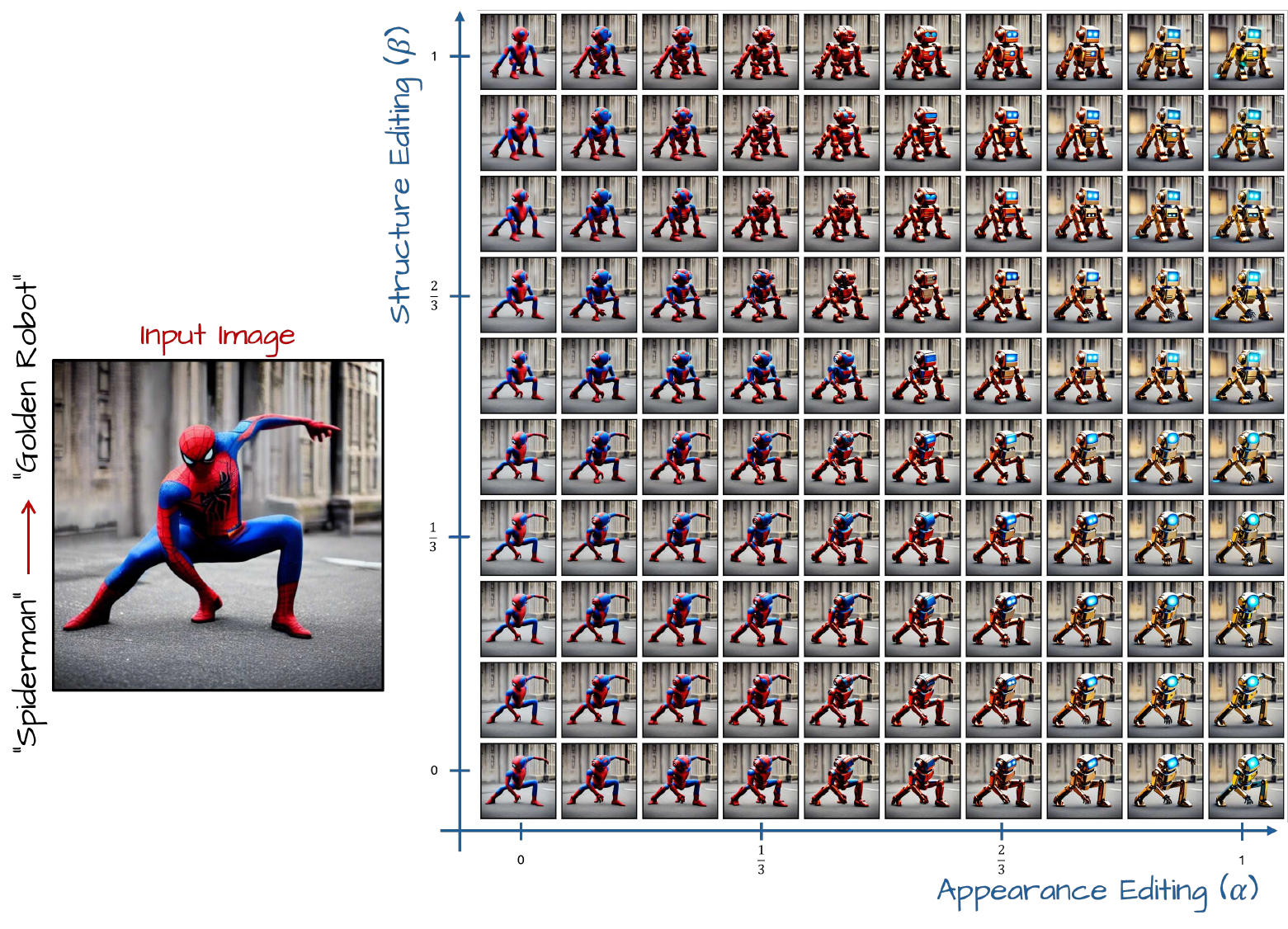}
    \caption{Additional results showcasing our correspondence‐aware attention interpolation and structural alignment. Adjusting $\alpha$ smoothly shifts the appearance from the source to the target, while varying $\beta$ progressively alters structural elements. The grid shows how appearance and structure can be controlled independently to achieve diverse transformations.}
    \label{fig:slider1}
\end{figure*}

\begin{figure*}[h]
    \centering
    \includegraphics[width=0.7\textwidth]{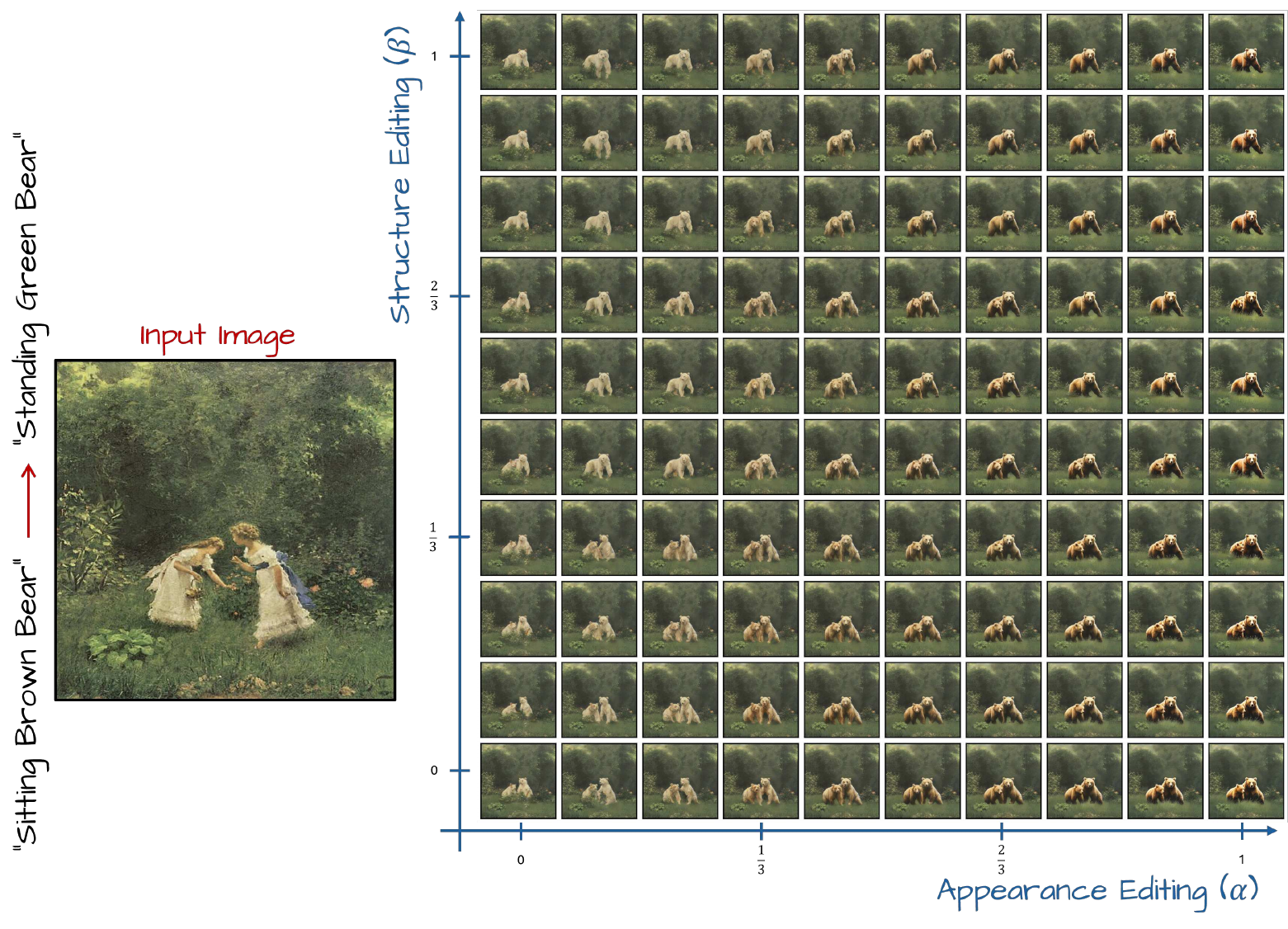}
    \caption{Additional results showcasing our correspondence‐aware attention interpolation and structural alignment. Adjusting $\alpha$ smoothly shifts the appearance from the source to the target, while varying $\beta$ progressively alters structural elements. The grid shows how appearance and structure can be controlled independently to achieve diverse transformations.}
    \label{fig:slider2}
\end{figure*}

\begin{figure*}[h]
    \centering
    \includegraphics[width=0.7\textwidth]{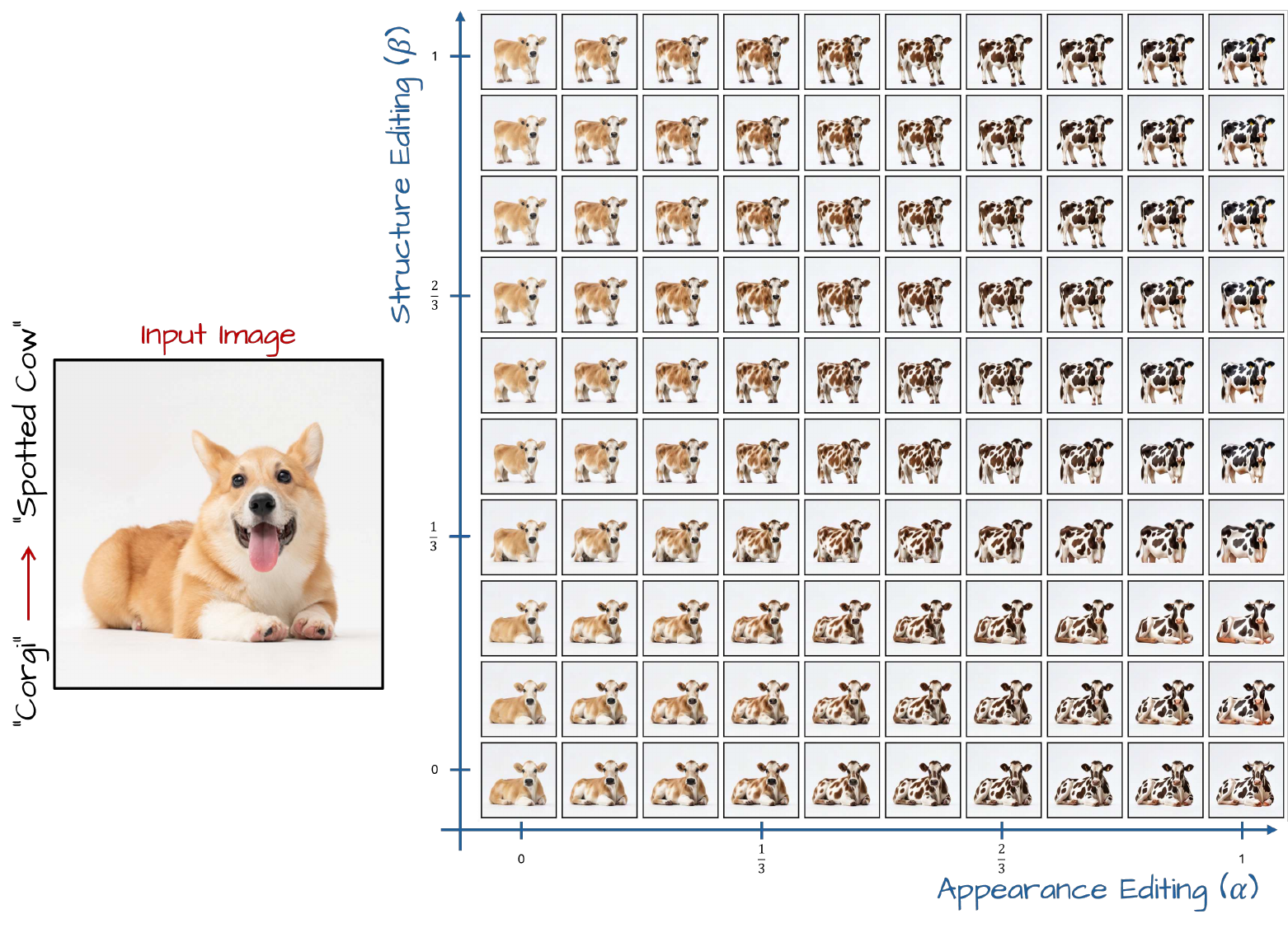}
    \caption{Additional results showcasing our correspondence‐aware attention interpolation and structural alignment. Adjusting $\alpha$ smoothly shifts the appearance from the source to the target, while varying $\beta$ progressively alters structural elements. The grid shows how appearance and structure can be controlled independently to achieve diverse transformations.}
    \label{fig:slider3}
\end{figure*}

\begin{figure*}[h]
    \centering
    \includegraphics[width=0.7\textwidth]{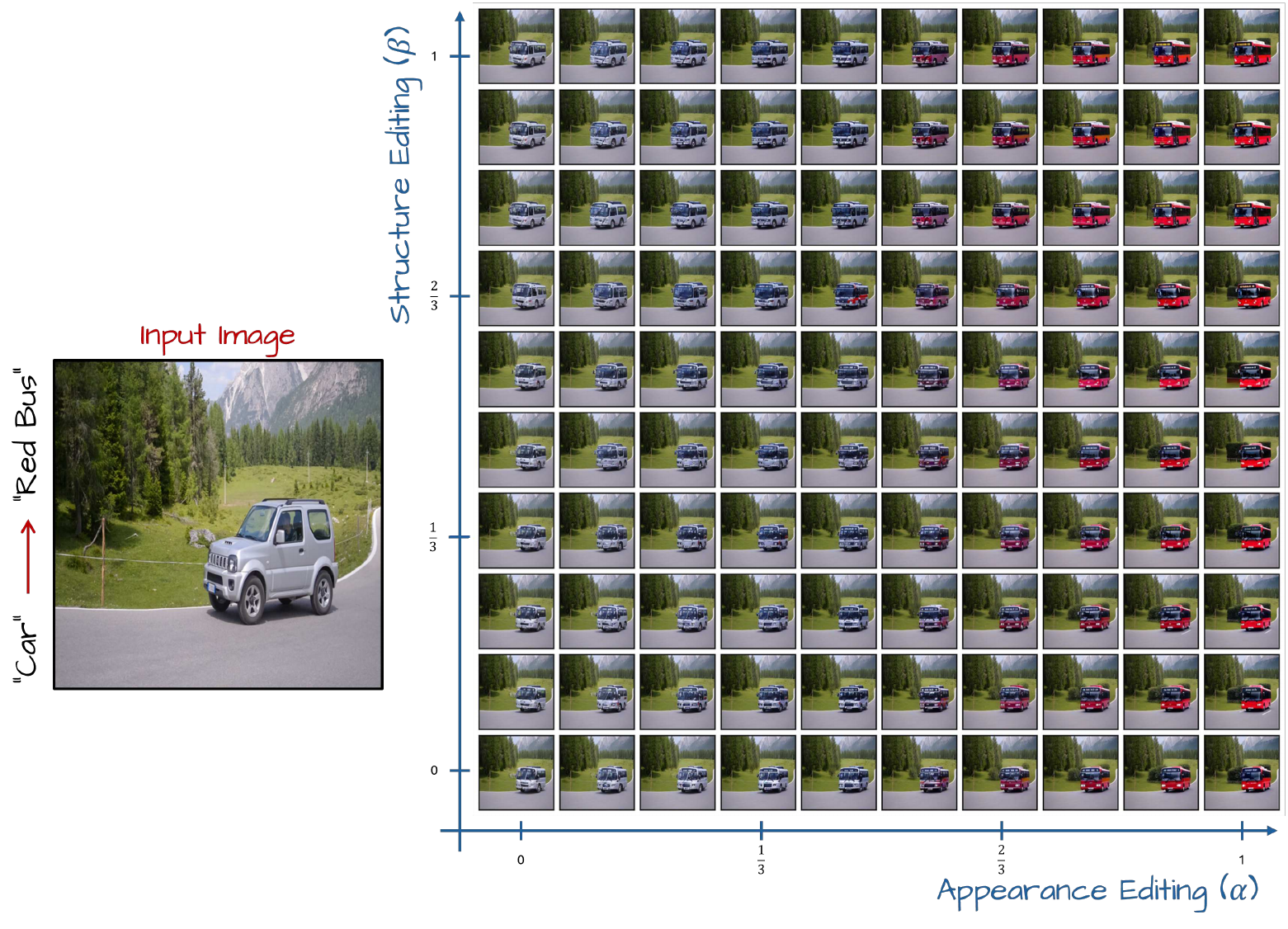}
    \caption{Additional results showcasing our correspondence‐aware attention interpolation and structural alignment. Adjusting $\alpha$ smoothly shifts the appearance from the source to the target, while varying $\beta$ progressively alters structural elements. The grid shows how appearance and structure can be controlled independently to achieve diverse transformations.}
    \label{fig:slider4}

\end{figure*}

\begin{figure*}[h]

    \centering
    \includegraphics[width=0.7\textwidth]{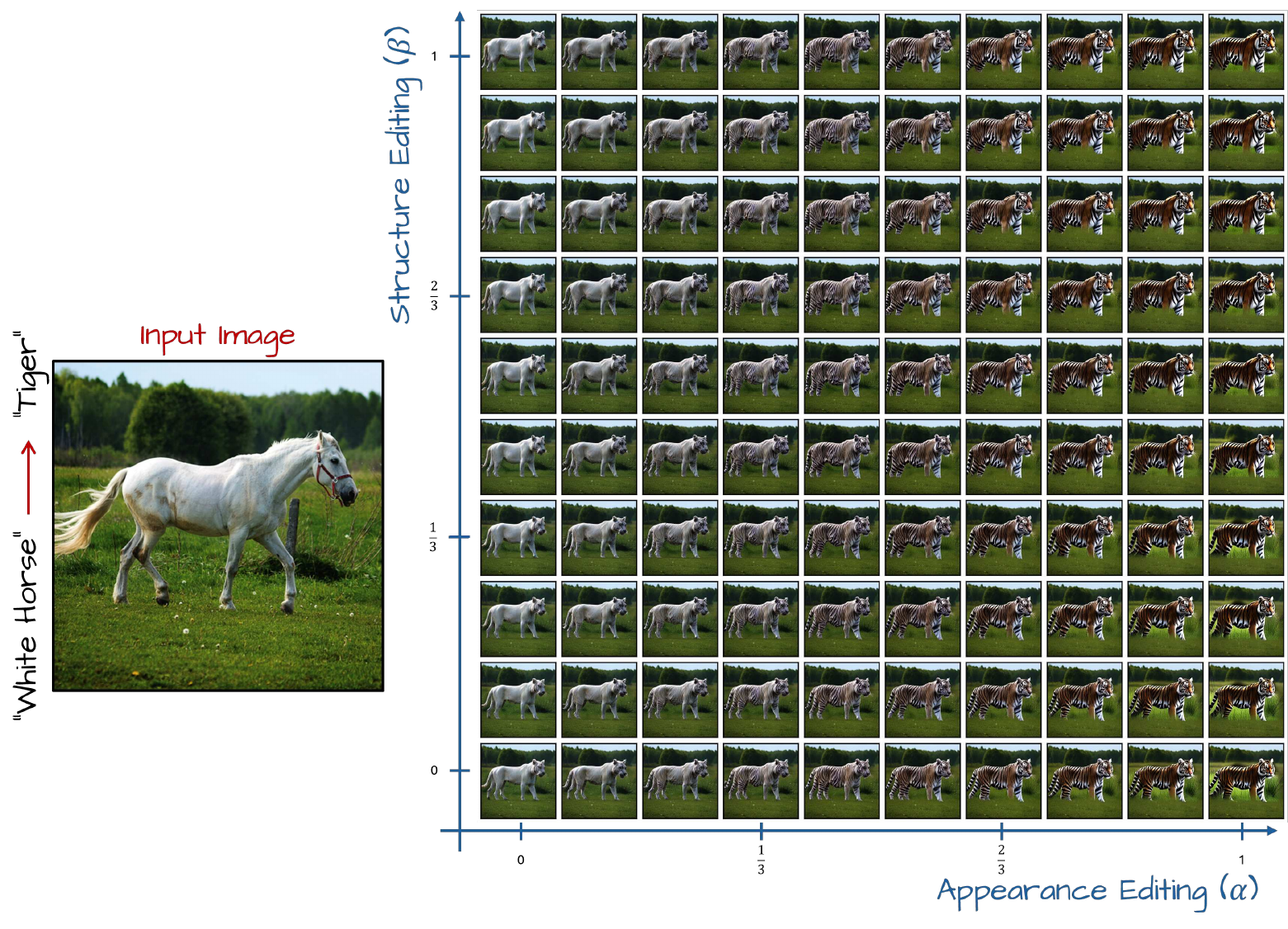}
    \caption{Additional results showcasing our correspondence‐aware attention interpolation and structural alignment. Adjusting $\alpha$ smoothly shifts the appearance from the source to the target, while varying $\beta$ progressively alters structural elements. The grid shows how appearance and structure can be controlled independently to achieve diverse transformations.}
    \label{fig:slider5}
\end{figure*}

\begin{figure*}[h]
    \centering
    \includegraphics[width=0.7\textwidth]{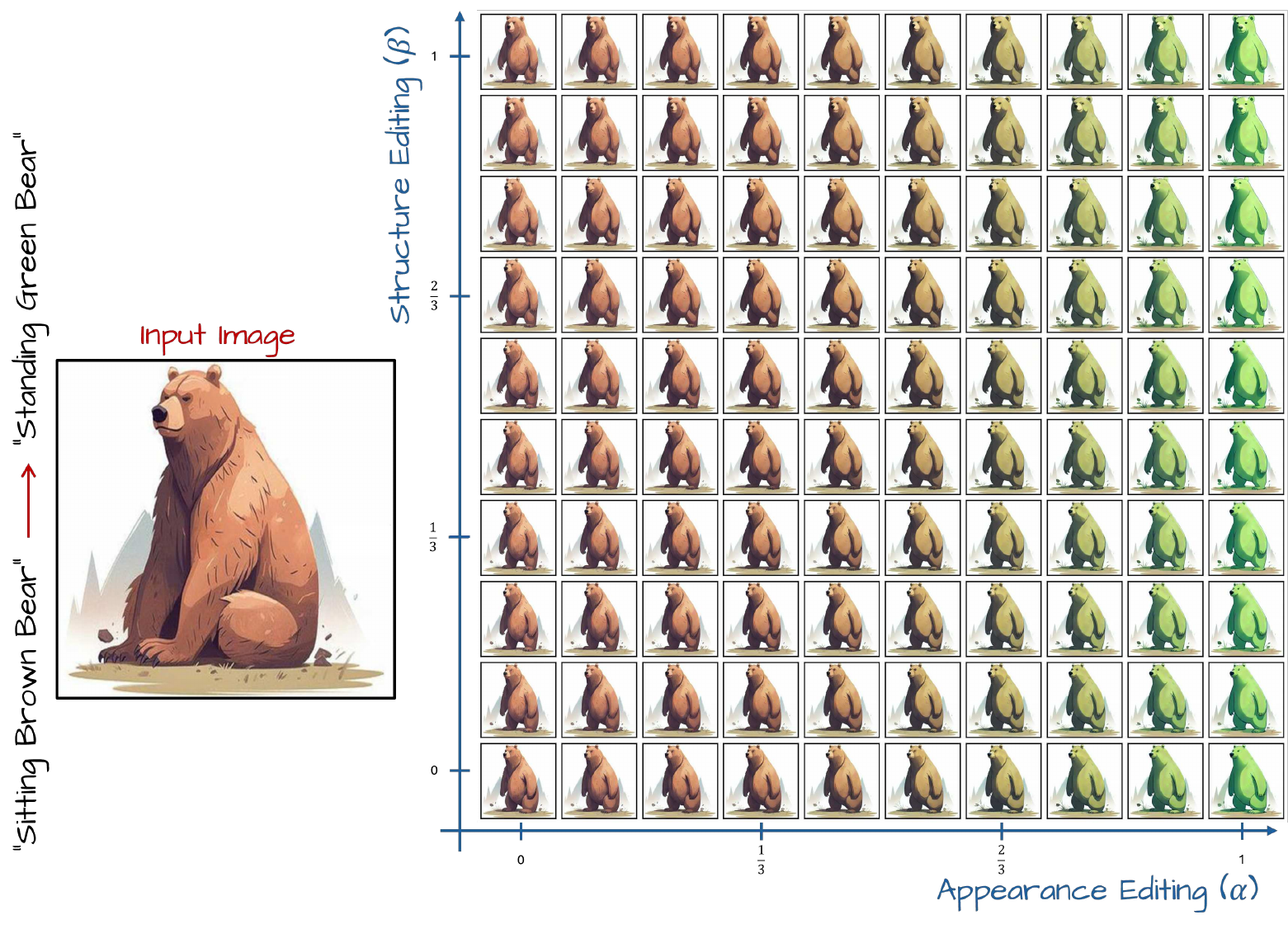}
    \caption{Additional results showcasing our correspondence‐aware attention interpolation and structural alignment. Adjusting $\alpha$ smoothly shifts the appearance from the source to the target, while varying $\beta$ progressively alters structural elements. The grid shows how appearance and structure can be controlled independently to achieve diverse transformations.}
    \label{fig:slider6}
\end{figure*}

When edits rely solely on text instructions, users often find it hard to specify how much of the source’s appearance or structure should remain intact. We address this by introducing two user-defined parameters, $\alpha$ and $\beta$, and, because our approach only uses a few denoising steps, users can quickly adjust $\alpha$ and $\beta$ to reach the desired outcome with minimal time overhead. As shown in Figure~\ref{fig:slider1} to Figure~\ref{fig:slider6}, tuning $\alpha$ determines how much of the original look is preserved versus newly generated, while $\beta$ refines how strictly the layout follows the source. These parameters provide clear, fine-grained control, reducing guesswork and enabling a broad range of edits, from slight refinements to large-scale transformations, while preserving stable and predictable outcomes.

\section{Latent Correction via Patch-based Correspondence}
\label{sec:latent_correction}

\subsection{Why the Original Corrections Fail}
Noise–inversion of the input image returns a sequence of latent
corrections $\{z_t\}_{t=1}^T$ such that, if we inject the same $z_t$’s during the backward pass and keep the conditioning text unchanged, the model reconstructs the source image pixel‑perfectly.
Each $z_t$ therefore assumes that every object will still occupy the exact spatial location it had in the source.

Editing, however, breaks this assumption: for instance, if a dog is asked to jump, the pixels affected by the edit are now displaced, and no longer correspond to the locations for which the original $z_t$ corrections were computed. Reusing those unmodified corrections forces the model to inject noise that is no longer spatially aligned, leading to mismatched textures, silhouette glitches, and the reappearance of content that no longer belongs. 

\subsection{Solution: Correspondence‑aware Latent Corrections} 
\begin{figure*}[h]
    \centering
    \includegraphics[width=0.76\textwidth]{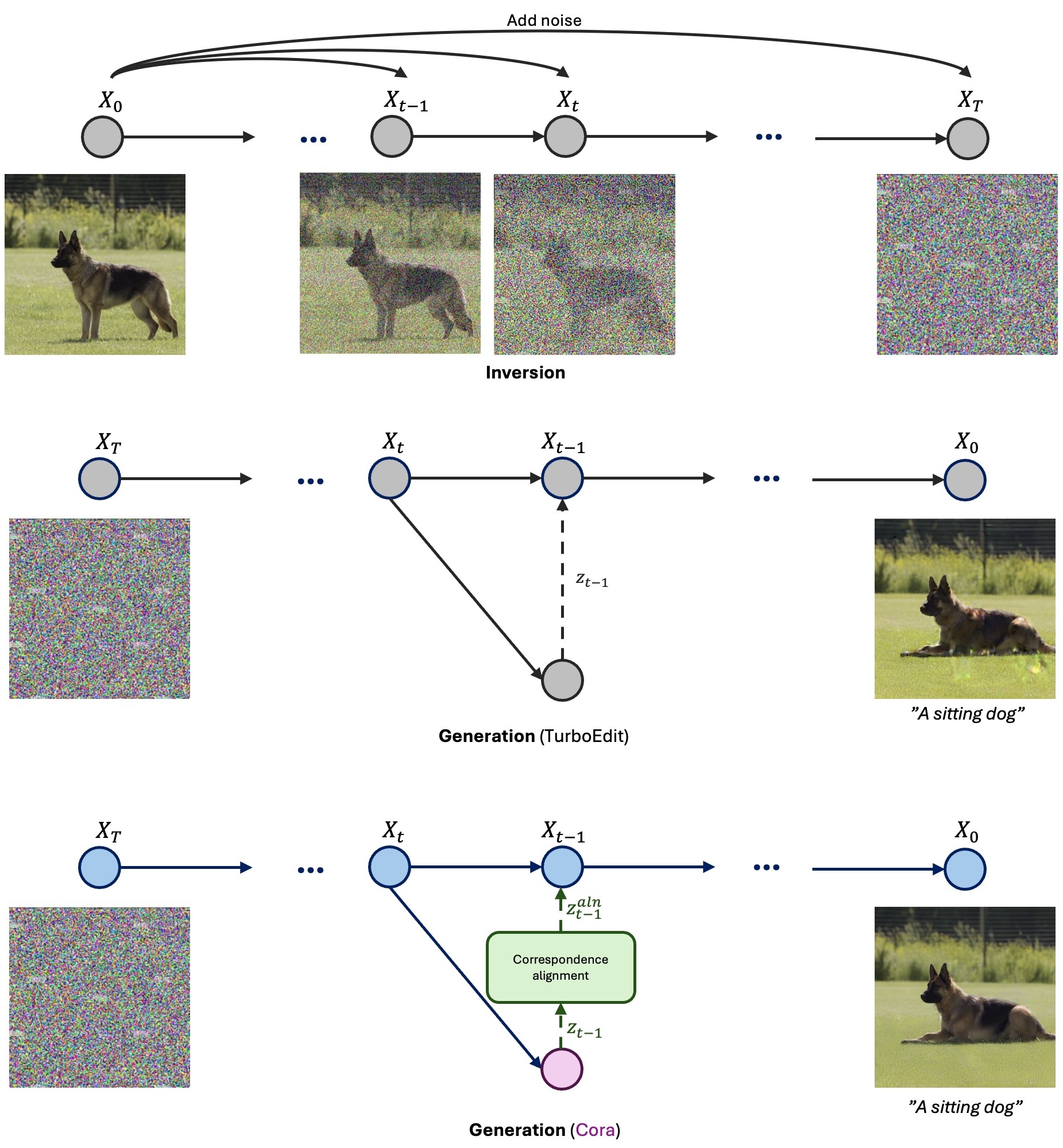}
    \caption{\textbf{Correspondence-aware latent correction.} 
    Top: inversion turns the source image into noise \(x_T\) and extracts residuals \(\{z_t\}\) that exactly reconstruct the original pose. 
    Middle: TurboEdit re-injects the unaligned residual (correction term) \(z_{t-1}\) while editing the pose, causing textures to snap back to old positions. 
    Bottom: our method aligns \(z_{t-1}\) via DIFT-based patch correspondences, producing a geometry-aware correction \(z^{\text{aln}}_{t-1}\) and a clean, artifact-free result.}
    \label{fig:supp_latent_correction}
\end{figure*}

Fig.~\ref{fig:supp_latent_correction} contrasts the standard TurboEdit pipeline (middle) with our correspondence‑aware variant (bottom). After inversion (top) we still obtain the original noise residual $z_{t-1}$, but before re‑injecting it we \emph{realign} it so that it follows the geometry of the edited frame $x_{t-1}$.

Our patch-based correspondence approach divides DIFT feature maps into overlapping patches, matches each target patch to its most similar source patch, and then reconstructs the aligned target representation by reassembling the matched patches back into their original spatial locations and averaging overlapping regions.

\subsection{Patch Extraction and Matching}

Let $D_S \in \mathbb{R}^{C \times H \times W}$ and $D_T \in \mathbb{R}^{C \times H \times W}$ be the
feature maps of the source and target images, respectively. We extract
overlapping patches of size $k \times k$ from both maps with stride $s$. This
turns each map into a set of flattened patch vectors:
\begin{equation}
\begin{aligned}
\mathrm{patches}_S &= \{ p_{S,1}, \dots, p_{S,N_S} \}, \quad p_{S,i} \in \mathbb{R}^{Ck^2}, \\
\mathrm{patches}_T &= \{ p_{T,1}, \dots, p_{T,N_T} \}, \quad p_{T,j} \in \mathbb{R}^{Ck^2}.
\end{aligned}
\end{equation}

Here, $N_S$ and $N_T$ denote the total numbers of patches extracted from the
source and target, respectively.

We then compute cosine similarity between every pair $(p_{T,j}, p_{S,i})$:
\begin{equation}
  \mathrm{sim}\bigl(p_{T,j}, p_{S,i}\bigr) \;=\;
  \frac{p_{T,j} \cdot p_{S,i}}{\|p_{T,j}\|\,\|p_{S,i}\|}.
\end{equation}
For each target patch $p_{T,j}$, we select the index of the best matching source
patch:
\begin{equation}
  \label{eq:bestmatch}
  C_{T\rightarrow S}(j) \;=\; \arg\max_{1 \le i \le N_S}
  \mathrm{sim}\bigl(p_{T,j}, \, p_{S,i}\bigr).
\end{equation}

\subsection{Reassembling Matched Patches}
\label{subsec:patch_reassemble}

After finding the best-match index for each target patch, we replace each target
patch $p_{T,j}$ with its matched source patch $p_{S,\mathcal{I}(j)}$, yielding
an updated patch set:
\begin{equation}
  \hat{p}_{T,j} \;=\; p_{S,\;C_{T\rightarrow S}(j)}.
\end{equation}
We then reassemble these patches into a tensor of shape $(C, H, W)$ by placing
each $\hat{p}_{T,j}$ at the corresponding spatial location of the $j$-th patch in
the target. Since patches overlap, every pixel can receive contributions from
multiple patches. Specifically, we denote by $\tilde{D}_T(\mathbf{u})$ the sum
of all matched-patch contributions at spatial location $\mathbf{u}$. Let
$\Omega_j$ be the set of pixel coordinates covered by the $j$-th patch. Then:
\begin{align}
  \tilde{D}_T(\mathbf{u}) &=
    \sum_{j: \;\mathbf{u} \in \Omega_j} \hat{p}_{T,j}(\mathbf{u}), 
  \\
  W(\mathbf{u}) &=
    \sum_{j: \;\mathbf{u} \in \Omega_j} 1,
\end{align}
where $\hat{p}_{T,j}(\mathbf{u})$ is the feature value of patch $\hat{p}_{T,j}$
for pixel $\mathbf{u}$. To normalize overlaps, we compute the final aligned map
$\hat{D}_T$ by dividing each spatial location by its total overlap:
\begin{equation}
  \hat{D}_T(\mathbf{u}) \;=\;
  \frac{\tilde{D}_T(\mathbf{u})}{\,W(\mathbf{u}) + \varepsilon\,},
\end{equation}
with $\varepsilon$ a small constant to prevent division by zero.

\section{Content-Adaptive Interpolation}

When a prompt leads to large deformations or introduces new objects, not every pixel in the edited (target) image should be forced to match a pixel in the original (source) image. Over-enforcing alignment often creates visual artifacts or incorrect texture transfers. To address this, we propose a two-step strategy that checks whether each target patch has a reliable counterpart in the source before blending.

\subsection{1. Bidirectional Matching}

Let $S$ be the set of source patches and $T$ be the set of target patches. For each source patch $s \in S$, we define its top-$k$ most similar target patches as:

\begin{equation}
\mathcal{K}(s) = \operatorname*{arg\,top_k}_{t \in T} \{ \text{sim}(s, t) \},
\end{equation}

where $\text{sim}(\cdot, \cdot)$ is the cosine similarity between feature vectors. Similarly, for each $t \in T$, define $\mathcal{K}(t)$. A pair $(s, t)$ is said to be \textit{bidirectionally matched} if $s \in \mathcal{K}(t)$ and $t \in \mathcal{K}(s)$. These are considered strong correspondences, and we blend source and target information using a user-defined weight $\alpha$.

\subsection{2. Weak Matches $\rightarrow$ Treat as New}

Some target patches remain unmatched. For each such target patch $t \in T$, we compute the strength of its best match in the source as:

\begin{equation}
\text{sim}_{\max}(t) = \max_{s \in S} \text{sim}(s, t).
\end{equation}

Let $U \subset T$ be the set of unmatched patches. We compute the $\gamma$-quantile threshold $\tau_\gamma$ over $\text{sim}_{\max}(t)$ values:

\begin{equation}
\tau_\gamma = \text{quantile}_\gamma\left( \{ \text{sim}_{\max}(t) \mid t \in U \} \right).
\end{equation}

If $\text{sim}_{\max}(t) < \tau_\gamma$, we classify $t$ as \textit{new} and set $\alpha = 1$, meaning it is fully guided by the prompt and not the source image.

\subsection{Practical Insights}

This content-adaptive interpolation balances preservation and generation: we reuse source appearance where reliable correspondences exist, and rely on prompt-driven generation in new regions. This avoids artifacts from over-aligning unrelated content. While the matching process is not perfectly accurate, we found it to be sufficient in most examples. In practice, it significantly reduces artifacts caused by mistakenly interpolating between unmatched patches.

Interestingly, even when only a portion of a new region, such as 20\% of the pixels corresponding to a newly generated object like a hat, is correctly identified as "new," the outcome is often satisfactory. Since we avoid interpolation for those pixels, they remain purely prompt-driven. During denoising, the rest of the image tends to adapt around these pixels, effectively completing the structure and appearance of the new content. Nonetheless, improving the accuracy of identifying new regions would further enhance the quality and reliability of the generated results.

\begin{figure*}[h]
    \centering
    \includegraphics[width=0.65\textwidth]{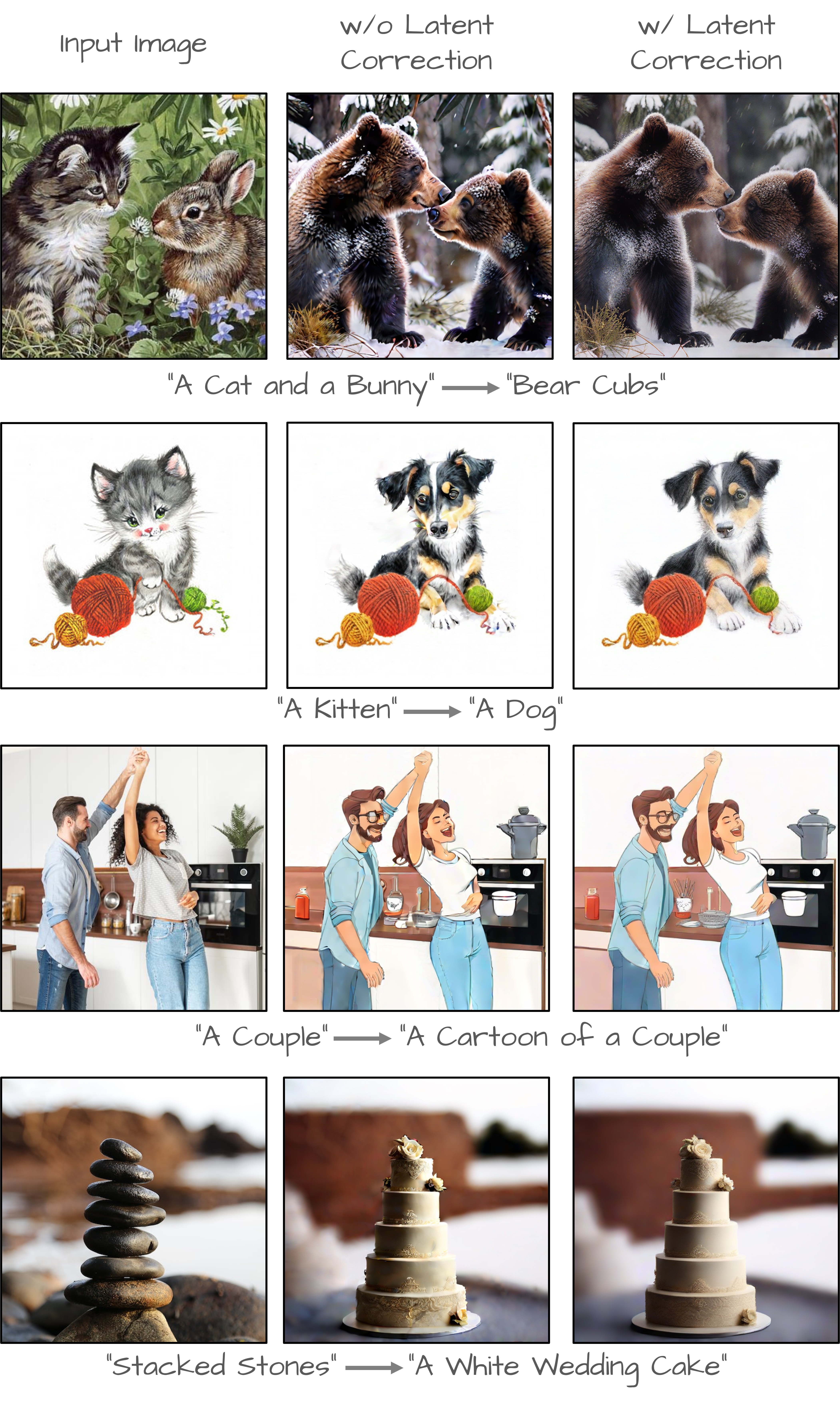}
    \caption{\textbf{Ablation on latent correction.} Without latent correction, multiple misalignment artifacts and unnatural deformations occur. Applying correction produces cleaner and more realistic results.}
    \label{fig:supp_correction2}
\end{figure*}


%% file: tables/quant_comp2.tex
\begin{table*}[t]
    \caption{Quantitative comparisons among text-based editing baselines. 
    \textbf{Bold} indicates the best scoring method, \underline{underline} indicates the second best, 
    \textcolor{blue}{blue} indicates the third best.}
    \label{tab:quant2}
    \vspace{-3pt}
    \renewcommand{\arraystretch}{1.2}
    \addtolength{\tabcolsep}{-0.31em}
    \footnotesize
    \resizebox{0.9\textwidth}{!}{
    \begin{tabular}{l | c c c c c c}
        & \shortstack{PSNR \\ (Background) \(\uparrow\)} 
        & \shortstack{LPIPS \\ (Background) \(\downarrow\)} 
        & \shortstack{MSE \\ (Background) \(\downarrow\)} 
        & \shortstack{SSIM \\ (Background) \(\uparrow\)} 
        & \shortstack{CLIP Sim. \\ (Whole) \(\uparrow\)}
        & \shortstack{CLIP Sim. \\ (Edited) \(\uparrow\)} \\

        \toprule
        Prompt2prompt~\cite{hertz2022prompt} & 19.9 & 153.98 & 188.94 & 79.58 & \textcolor{blue}{24.87} & \textcolor{blue}{22.35} \\
        Plug-and-play~\cite{tumanyan2022plugandplay} & 24.88 & 80.86 & 72.53 & 86.25 & \textbf{25.13} & 22.08 \\
        NTI~\cite{mokady2022null} & \textbf{29.92} & \textbf{36.68} & \textbf{25.18} & \textbf{91.02} & 24.18 & 21.09 \\
        StyleDiffusion~\cite{li2023stylediffusion} & \underline{28.65} & \textcolor{blue}{45.16} & \underline{34.53} & \textcolor{blue}{89.6} & 24.02 & 21.28 \\
        InstructPix2Pix~\cite{brooks2022instructpix2pix} & 22.47 & 145.50 & 241.69 & 80.38 & 21.54 & 19.80 \\
        DDPM inversion~\cite{HubermanSpiegelglas2023} & 24.09 & 110.68 & 220.50 & 84.65 & 23.35 & 20.40 \\
        MasaCtrl~\cite{Cao_2023_ICCV} & 24.47 & 82.81 & 78.44 & 87.01 & 23.99 & 21.00 \\
        TurboEdit~\cite{deutch2024turboedittextbasedimageediting} & 27.95 & \underline{44.98} & \textcolor{blue}{37.87} & \underline{89.82} & 24.79 & \underline{22.44} \\
        InfEdit~\cite{xu2023infedit} & 25.51 & 123.76 & 370.24 & 80.21 & 23.34 & 20.08 \\
        \paper (Ours) & \textcolor{blue}{28.02} & 50.57 & 38.03 & 89.39 & \underline{24.91} & \textbf{22.69} \\
        \bottomrule
    \end{tabular}
    }
    \vspace{-5pt}
\end{table*}